\documentclass[preprint]{elsarticle}

\usepackage{graphicx}
\usepackage{hyperref}
\usepackage[utf8]{inputenc}
\usepackage{amsmath}
\usepackage{amssymb}
\usepackage{subcaption}
\usepackage[ruled]{algorithm2e}
\DeclareMathOperator*{\argmax}{arg\,max}

\begin{document}

\begin{frontmatter}

\title{A near Pareto optimal approach to student-supervisor allocation with two sided preferences and workload balance}

%
\author[flo,cov,ui1]{Victor Sanchez-Anguix} \ead{vsanchez@florida-uni.es} \ead{ac0872@coventry.ac.uk} \ead{victor.sanchez.anguix@ui1.es}
\author[cov]{Rithin Chalumuri} \ead{chalumuriv@uni.coventry.ac.uk}
\author[ozy,tud]{Reyhan Aydo\u{g}an} \ead{reyhan.aydogan@ozyegin.edu.tr}
\author[upv]{Vicente Julian} \ead{vinglada@dsic.upv.es}


\address[flo]{Florida Universitaria, Carrer del Rei en Jaume I, 2, 46470, Catarroja, Valencia, Spain}
\address[ui1]{Universidad Isabel I, Calle de Fernán González 76, 09003, Burgos, Spain}
\address[cov]{Coventry University, School of Computing, Electronics and Mathematics, Gulson Rd, CV1 2JH, Coventry, United Kingdom}

\address[ozy]{\"{O}zye\u{g}in University, Department of Computer Science, Istanbul, Turkey}
\address[tud]{Delft University of Technology, Interactive Intelligence Group, Delft, The Netherlands}
\address[upv]{Universitat Politècnica de València, Departamento de Sistemas Informáticos y Computación, Camí de Vera s/n, 46022, Valencia, Spain}

\begin{abstract}

The problem of allocating students to supervisors for the development of a personal project or a dissertation is a crucial activity in the higher education environment, as it enables students to get feedback on their work from an expert and improve their personal, academic, and professional abilities. In this article, we propose a multi-objective and near Pareto optimal genetic algorithm for the allocation of students to supervisors. The allocation takes into consideration the students and supervisors' preferences on research/project topics, the lower and upper supervision quotas of supervisors, as well as the workload balance amongst supervisors. We introduce novel mutation and crossover operators for the student-supervisor allocation problem. The experiments carried out show that the components of the genetic algorithm are more apt for the problem than classic components, and that the genetic algorithm is capable of producing allocations that are near Pareto optimal in a reasonable time.
\end{abstract}

\begin{keyword}
genetic algorithms \sep student-project allocation \sep matching \sep Pareto optimal \sep artificial intelligence
\end{keyword}

\end{frontmatter}

\section{Introduction}

Every year in higher education (HE) institutions, students undertake individual projects that are supervised by a tutor that offers academic advice and guidance, either as an undergraduate or master dissertation, as part of their coursework, or simply as a summer research project.  Students are usually allocated to supervisors for their projects by means of a centralized human decision maker or by means of interactions between students and staff members. The decision makers have to take into consideration the preferences of both students and supervisors with respect to the conduct of the project, as well as departmental constraints such as minimum and maximum levels of workload (in terms of supervision) for each supervisor. This situation results in an extremely time consuming process, and a suboptimal allocation due to a large and complex search space faced by human decision makers. Automating this process by applying artificial intelligence techniques may enhance the process in terms of satisfaction and performance of students with these individual projects.

In this article, we present a genetic algorithm for matching students to supervisors according to both students' and supervisors' preferences and the constraints of the department. The rationale behind this problem is matching an appropriate student with a supervisor for the development of an individual project. The problem of matching students to supervisors, or students to projects \cite{anwar2003student, harper05, abraham2007two, manlove2008student, srinivasan2008efficient, pan2009multi, el2012artificial,  iwama2012improved, kwanashie2014profile, kwanashie2015efficient, salami2016genetic, cooper20183, chiarandini2018handling}, is a subclass of the wider problem of matching between two sets, one of the most studied fields in computer science due to its applications to a wide range of domains  such as the hospital/residents (HR) or the college admission (CA) problem \cite{biro2010college, hamada2011hospitals, hamada2016hospitals}. Particularly, the student-supervisor allocation problem solved in this article can be considered as an instance of the CA problem with lower and upper quotas, where the colleges are the supervisors, both colleges and students (i.e., supervisors and students in our case) have some representation of preferences on each other for the conduct of a project, and the minimum and maximum quotas are the minimum and maximum number of students to be supervised by staff members. In this situation, it has been shown that there is no guarantee for a stable allocation to exist \footnote{Stability defined as the lack of incentive for any pair of student and college to change their current allocation in favor of one that allocates them together.} and even looking for a near-stable allocation is a NP-hard complex problem \cite{hamada2011hospitals,hamada2016hospitals}.

In order to tackle the complexity mentioned above, in this article we propose a multi-objective and Pareto optimal genetic algorithm (GA). The main highlights of the GA proposed in this article are: (i) it takes into consideration both the preferences of the students and the supervisors with respect to type of project to undertake/supervise; (ii) it considers the constraints on the individual minimum and maximum supervision workload of each supervisor; (iii) it aims to provide a fair and balanced allocation in terms of the workload for each supervisor; (iv) it provides multiple near-optimal solutions considering both the students' preferences and the preferences of the supervisors; (v) it provides multiple near Pareto optimal solutions that can be used by decision makers to trade-off between the multiple objectives optimized; (vi) and the GA employs novel mutation and crossover operators in the context of the student-supervisor allocation problem, specifically designed for allocation problems with lower and upper bound supervision quotas. 

The algorithm has been tested using real preferences elicited from students and supervisors, and it has been compared with classic GA components, with the proposed operators outperforming classic ones for the problem at hand. The rest of the article is organized as follows. First, we highlight the differences between the problem tackled in this article and the work carried out by relevant studies in Section \ref{ref:related}. Then, we elaborate on problem description in Section \ref{ref:prob}. Once the problem has been formally defined, the proposed genetic algorithm for student-supervisor allocation is explained in Section \ref{ref:GA}. Section \ref{ref:experiments} provides empirical evaluation of the proposed approach. Lastly, we conclude our work with future work directions in  Section \ref{ref:conclusions}. 


\section{Related work}
\label{ref:related}

The problem of allocating the students to supervisor is a one-to-many matching where we allocate only one supervisor to each student while more than one student can be assigned to a supervisor. As mentioned above, this particularity makes it similar to the college admission (CA) and the hospital/residents (HR) problem, two well-known one-to-many matching problems from the point of view of theoretical computer science. In the HR problem, each resident has a ranked list of preferences on the hospitals they may be assigned to, and hospitals also have ranked preferences on the residents they may accept. Similarly, in the CA problem, each student has ranked preferences on the colleges they may be accepted, and each college has ranked preferences on the students that they may accept. Both colleges and hospitals may accept more than one student/resident, making it a one-to-many matching. 

Our student-supervisor matching problem involves a one-to-many matching where supervisors have both lower and upper supervision quotas, and both sides have preferences on each other. Biro \textit{et al.} \cite{biro2010college} studied the problem of the CA problem with both lower and upper quotas from a theoretical perspective. Differently to our setting, the work presented in \cite{biro2010college} allows for colleges to be closed in case that their minimum acceptance quota is not reached. The authors found that, in the presence of both lower and upper quotas, there may not exist a stable matching. A matching is considered stable if for every pair of student and college not included in the matching, either the student is matched to a college that he/she prefers, or the college quota is full with applicants that the university prefers. In addition to this, determining if a stable matching exists in this setting is a NP problem. Biro \textit{et al.} showed that a student oriented polynomial algorithm and a college oriented polynomial algorithm can be provided in the case that colleges are organized in nested sets and have common quotas. In our work, the minimum supervision quota for each supervisor must be achieved, and all specified staff members participate in the allocation.

Later on, Hamada \textit{et al.} \cite{hamada2011hospitals,hamada2016hospitals} further studied the problem of matching with lower and upper quotas, focusing on scenarios where all colleges/hospitals should reach their minimum quota in the allocation. In this particular scenario, it is proved that the problem of providing a matching that is as close as possible to be stable is still a NP-hard problem, but a polynomial time algorithm approximation algorithm exists with an approximation guarantee equal to the sum of the number of hospitals and residents. This approximation may not be appropriate for large numbers of hospitals and residents or student and supervisors, as the problem faced in our proposal.

The previous findings provide the reader with some background on the complexity of the matching problem presented in this article. This complexity, and the need to appropriately tackle large problems motivated the choice of a metaheuristic instead of a global optimization technique. In the next few lines, we discuss how our present work compares to other student-project or student-supervisor allocation schemes proposed in the literature.

Anwar \textit{et al.} \cite{anwar2003student} were one of the pioneering authors in providing a computational solution to the student-project allocation problem. The article introduces two different integer programming models: one to allocate students to projects while minimizing the projects supervised by staff members, and another to maximize the students' satisfaction according to their preferences on group projects to be allocated and to be undertaken. In this setting, staff members propose a list of projects and students provide a rank of four projects to be allocated on. Both integer programming models were tested on a real dataset consisting of 60 projects, 22 staff members, and 39 students. Similarly, \cite{harper05} introduces the use of genetic algorithms for solving the student-project allocation problem. In their setting, students provide a ranked list with their most preferred projects, and each student is allocated a project from the provided list, with projects being carried out individually. The algorithm was tested with real data consisting of 25 students and 34 projects, and also with problems created from data provided from the OR-library \cite{beasley1990or}. These models only take into consideration the students preferences, but they do not consider the staff preferences with regards to projects and students and the workload of supervisors. In addition to this, they can only optimize a single objective function which precludes decision makers from trading-off between the students' and the staff preferences.

Abraham \textit{et al.} \cite{abraham2007two} focus on solving the student-project allocation problem from an optimal perspective. The authors assume that a list of projects is provided by staff members. The students provide a ranked list of their most preferred projects, while staff members explicitly rank students that desire to be allocated to the staff member's projects. Under this assumption, the authors provide two linear algorithms to find stable matching: one from the perspective of the students' preferences, and another from the perspective of the lecturers' preferences. While an optimal solution can be guaranteed employing these algorithms, they either provide the optimal solution for the students or the optimal solution for the staff members, but no trade-off opportunity is provided to decision makers. In addition to this, the algorithms do not take into consideration the workload of supervisors, with the possibility of producing unbalanced solutions. Finally, it should also be considered that supervisors explicitly rank students which may not be feasible if supervisors do not know students, or it may be unfair for students with lower marks as many will end up in the last rank positions in lecturers' preferences.

Later on, Manlove \& O'Malley \cite{manlove2008student} study the student-project allocation problem in a scenario where students and supervisors have preferences over a set of projects. Both projects and supervisors have capacity constraints. Under these conditions, the authors prove that stable matchings can have different cardinalities, and thus the objective is that of finding the stable matching with a maximum cardinality. Solving this problem is NP hard, but the authors provide a student oriented approximation algorithm with a performance guarantee of 2 (i.e., only guaranteeing half of the cardinality of the maximum stable matching) and polynomial complexity. Iwama \textit{et al.} \cite{iwama2012improved} further narrowed down this bound to a range between 1.5 and 1.10. The proposed algorithms focus on optimizing the students' preferences, with no explicit consideration of the staff members' preferences, the workload of supervisors, or lower quota constraints.

Another genetic approach to the student-project allocation problem was provided by Srinivasan \& Rachmawati \cite{srinivasan2008efficient}. The described scenario consists of students providing a ranked list of projects from a list published every year by lecturers. The problem is tackled as multiobjective optimization problem where both the preferences of the students and the departments are taken into consideration. In order to compute the preferences of the department, the academic performance of the students and the workload of supervisors/departments are taken into consideration. As mentioned, assigning projects on merit may lead to undesirable situations whereby low performing students end up in less attractive projects. In addition to this, the model does not support lower and maximum supervision quotas for lecturers. Finally, it should be highlighted that despite the fact that multiple objectives are considered, these are aggregated into a single and final objective function. This requires to compute the GA every time that the human decision maker desires to trade-off between different objectives.

The work presented on \cite{pan2009multi} proposes the use of goal programming to tackle the student-project allocation as a hierarchical multiobjective problem.  The maximum priority of the model is maximizing the number of allocated students, and then it attempts to maximize the students' preferences and then the academic performance of allocated students. Again, the model employs academic performance to prioritize the departments' choice, which may be discriminatory. Moreover, the model does not allow to execute trade-offs between the different objectives, and it does not guarantee any degree of optimality for each of them.

The authors in \cite{el2012artificial} present an artificial immune system optimization algorithm for the student-project allocation problem. More specifically, the authors model a problem where a set of students and projects exist, and students have preferences on the projects to undertake. In their framework, students must be matched a project, and a project can be matched at most once. The authors study the performance of several mutation operators on the problem, although they focus on swapping projects between students based on different criteria like time. As it will be appreciated, our proposed mutation operator takes into consideration both swapping (students between supervisors) and transferring operations (giving a student to another supervisor) and they consider the minimum and maximum supervision quota of each supervisor.

In \cite{kwanashie2014profile}, the authors focus on solving the student-project allocation problem where only students' preferences are present, but supervisors have both lower and upper supervision quotas. In the article, the authors provide efficient algorithms that aim to provide optimal solutions in the context of a single side optimization (i.e., students' preferences). For that, their proposed algorithms guarantee finding greedy maximum matchings or generous maximum matchings. The first aims to find the largest matching in terms of the number of students allocated, and maximizing the number of students allocated their first and most preferred choices. The second aims the largest matching that minimizes the number of students allocated their least preferred choices. The work presented by the authors does not support lecturers' preferences, and thus optimizes a single objective criteria, and matchings found do not necessarily guarantee matching all of the students to projects/supervisors, something that we have considered fundamental in our present work.

Salami and Mamman propose another genetic algorithm for scenarios where students have complete preferences on supervisors, and supervisors have a maximum supervision quota~\cite{salami2016genetic}. However,  there is no consideration on supervisors' preferences or the workload balance for supervisors.

In \cite{chiarandini2018handling} the authors present mixed integer programming models for solving the student-project allocation problem with one-sided preferences (i.e., students). Differently to other approaches, students apply for projects in teams and the maximum capacity of projects is defined in number of teams rather than the number of individuals. The main focus of the article consists of analyzing different fairness metrics from the point of view of the students' allocation.

Recently, Cooper and Manlove \cite{cooper20183} have revisited the problem of allocating students to projects, where both students and lecturers have preferences over each other, and lecturers and projects have upper capacity constraints. The authors have provided a 3/2-approximation algorithm capable of calculating maximum stable matchings in linear time. It should be considered that this work does not include lower quotas and neither introduces fair balancing for the supervisors' workload. 

Our approach is based on students' and supervisors' preferences on project topics rather than projects. This is an advantage as it does not require lecturers to propose projects prior to the allocation and they can be negotiated with students according to their research interests. Furthermore, it does not discriminate students according to their performance as staff preferences' are based on topics rather than students. Most analyzed works only take into consideration the students' preferences \cite{anwar2003student, harper05, el2012artificial, kwanashie2014profile, salami2016genetic, chiarandini2018handling}, or they base the department preferences on pure academic merit/opinion \cite{abraham2007two, srinivasan2008efficient, pan2009multi}. We consider both the students' and the departments' preferences by adopting a multiobjective approach that provides decision makers with flexibility to trade-off between objectives as it estimates Pareto optimal solutions. The works analyzed in this section are either single objective (sided) \cite{anwar2003student,harper05,manlove2008student,el2012artificial, kwanashie2014profile, salami2016genetic, chiarandini2018handling} or they adopt a multiobjective stance by aggregating or prioritizing objectives \cite{srinivasan2008efficient,pan2009multi} or by focusing on finding efficient matchings with no lower quotas \cite{manlove2008student, iwama2012improved, cooper20183}. In addition to this, we aim to provide a balanced allocation that takes into consideration the workload of lecturers, a characteristic that is only present in \cite{srinivasan2008efficient}. Therefore, the proposed model should be more apt for the student-project allocation problem described in this article.

\section{Problem definition}
\label{ref:prob}
In this section we describe the problem of allocating students to supervisors from a formal perspective.  
Let $\mathcal{S}=\{s_{1},\dots,s_{n}\}$ and $\mathcal{R}=\{r_{1},\dots,r_{m}\}$ represent a set of students and a set of supervisors where $n$ and $m$ denote the number of students and number of supervisors respectively. 

\subsection{Matching definition}

A matching\footnote{Please note that the definition of matching employed in this paper aligns with the definition employed in the student-project allocation problem. It should be highlighted that this is different to the classic definition of matching in a graph.} $M$ is an assignment of students to supervisors, where each student is assigned exactly one supervisor. Without loss of generality, we say that $M(s_{i})$ represents the supervisor assigned to student $s_i$ in $M$, and $M(r_j)$ represents the set of students assigned to supervisor $r_j$ in $M$. Each supervisor $r_j$ has an upper bound supervision quota $c_{j,max}$, which is normally established by the head of the department or school. Similarly, each supervisor $r_j$ has a lower supervision quota $c_{j,min}$, set by the department or school, that determines the minimum number of students that he/she should supervise. This is the case in many higher education institutions, where supervisors have different teaching loads and therefore, they may be more or less available to supervise students' projects.

Given a matching, we say that a supervisor $r_j$ is under-subscribed iff $ c_{j,min} \leq |M(r_j)| < c_{j,max}$, he/she is full iff $|M(r_j)| = c_{j,max}$, and he/she is over-subscribed iff $|M(r_j)| > c_{j,max}$. We say that a matching $M$ is feasible iff $\forall r_j \in \mathcal{R}, c_{j,min} \leq |M(r_j)| \leq c_{j,max}$, i.e. for every supervisor he/she is full or under-subscribed. 

\subsection{Workload definition}
As mentioned in the article, we aim to consider the balance of the workload for supervisors when constructing a proper allocation with our GA. Therefore, we must provide a formal definition for what we consider workload and how to measure the balance of the workload in a matching $M$. The workload level of a supervisor $r_j$ in $M$ as $l_{j}=\frac{|M(r_j)|}{c_{j,max}}$. Namely, that is the ratio of students supervised in the matching $M$ over the maximum number of students that can be supervised by $r_j$. Analogously, we can define and $\mathcal{L}_{M}=\{l_1,\dots,l_m\}$ as a vector that contains the workload levels for all supervisors in the matching $M$, and we define $\sigma_{\mathcal{L}_M}$ as the standard deviation of the workload levels of supervisors in the vector $\mathcal{L}_{M}$.

\subsection{Evaluation of a student-supervisor assignment}
\label{ref:pref}
The first step towards evaluating the quality of a matching $M$ is that of evaluating the individual allocation of student $s_i$ to supervisor $r_j$. We define $V_{i,j}$, as the value given by a student $s_i$ to being allocated a supervisor $r_j$, and $V'_{j,i}$ as  the value given by a supervisor $r_j$ to being allocated a student $s_i$. 

In this work we assume that students cannot explicitly provide a complete list of preferences for supervisors. Even if they could provide a partial list of supervisors in rank of preference, the list would be biased to the supervisors that they like or the ones that they have met. It is not possible for the students to know all of the staff members in a relatively large school or department. There are different reasons for this. Additionally, students may hesitate to specify their preferences on their supervisors/teachers directly due to academic reasons and privacy issues. However, it is easy for them to specify which topics they would like to work on more. Therefore, we consider that students are able to provide a ranked list of $k$ topics to represent their preferences in the problem. Again, we assume that only $k$ explicit preferences can be given as the number of potential topics may be extremely large in some areas and that may result in a too costly elicitation process. Similarly, we can state the same about supervisors. They cannot explicitly rank all of the students as they may not know them. In addition to this, by making supervisors express their preferences in terms of project topics rather than students, we avoid discrimination according to academic performance. 

\begin{figure}
\includegraphics[width=\linewidth]{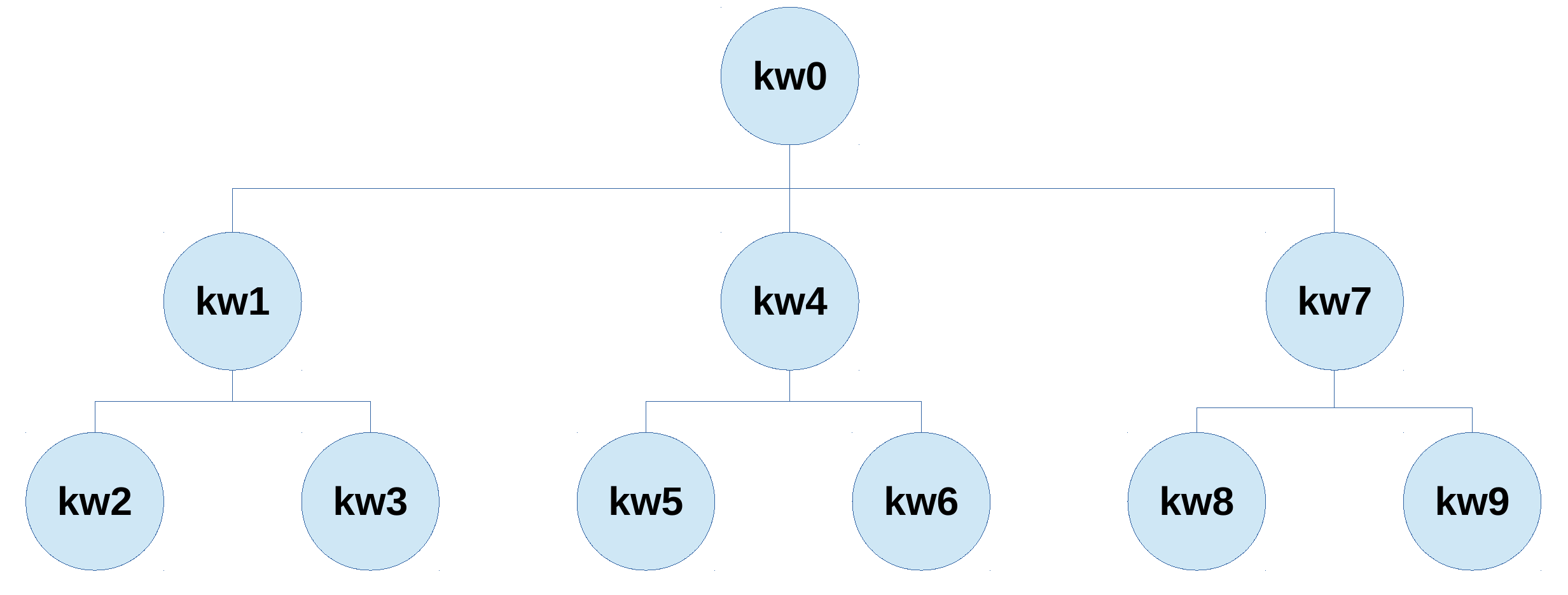}
\caption{Topics organized in a tree-like structure}
\label{img:tree}
\end{figure}

In order to evaluate both the students' and supervisors' preferences on topics, we assume that the topics are represented in a tree-like and hierarchical structure with a common root. In this structure, topics may be further divided into subtopics and so on, but it is always possible to relate how similar two topics are by analyzing the tree structure~\cite{Aydogan-2007}. An example of this tree-like structure can be observed in Figure \ref{img:tree}. In the example, let us assume that a student has stated that his preferred topic is $kw5$. Given the tree structure, one can easily compare how similar or related other keywords are based on the the number of common nodes in the tree structure. This is the case, despite the fact that the student may have not explicitly provided preferences for other keywords. We assume that each supervisor/student preferences are represented by a list of $k$ different topics from a tree-like and hierarchical structure. A student or supervisor $i$ describes his preferences with a ranked list $\mathcal{KW}_i = \{kw_{1},\dots,kw_{k}\}$ of $k$ topics where $\forall j < k, kw_{j} \succ kw_{k}$. 

We consider that the similarity between a student's and a supervisor's preferences depends on two factors: the similarity of the keywords provided by both in their lists, and the position of those keywords in their ranked lists. First, we define the similarity between two keywords, and then we define the similarity between the positions occupied by two keywords in two ranked lists.
\begin{itemize}
	\item \textbf{Keyword similarity:} Let us consider that there is a tree defined by $\mathcal{T} = (\mathcal{KW},E)$ where $\mathcal{KW}=\{kw_1,\dots,kw_l\}$ is a group of $l$ different nodes that represent topics in an area of knowledge, and $E$ is a group of edges in the form $(kw_i,kw_j)$ indicating that $kw_j$ specializes the topic in $kw_i$. The similarity of $kw_j$ to $kw_i$ topics in $\mathcal{T}$ is defined as:
\begin{equation}
S_{\mathcal{T}}(kw_i,kw_j) = \frac{| path(kw_i,\mathcal{T}) \cap path(kw_j,\mathcal{T}) |}{|path(kw_i,\mathcal{T})|}
\end{equation}
where $path: \mathcal{KW} \times \mathcal{T} \rightarrow 2^{\mathcal{KW}}$ is a function that retrieves the path defined from the root of the tree $\mathcal{T}$ to the node $kw$ (included). As a consequence, we define the similarity of $kw_j$ to $kw_i$ as the number of common nodes in the path defined from the root to both topics. Please, the reader should bear in mind that this similarity metric is not symmetric, to consider the fact that more specific topics are only fully matched by topics of greater or the same specificity. Lastly, if we assume two lists of preferences $\mathcal{KW}_i$ and $\mathcal{KW}_j$, and a topic $kw_{i} \in \mathcal{KW}_i$, we define its best matching topic $kw^*_{i} \in \mathcal{KW}_j$ as $\underset{kw_{j} \in \mathcal{KW}_j}{\argmax} S_{\mathcal{T}}(kw_{i},kw_{j})$.
\item \textbf{Rank similarity:} We define $S_{rnk}$ as the rank similarity between two keywords in two ranked lists of preferences $\mathcal{KW}_i$ and $\mathcal{KW}_j$. This similarity metric represents the fact that the order of the topics in both the student and supervisor's preferences should matter as it denotes the degree of interest of expertise in the topic. For instance, let us assume that the topic \textit{artificial intelligence} is defined as the most preferred topic for a student, and the best matching keyword in a supervisor's list is \textit{machine learning}. However, this topic appears as the last in the list of preferences for the supervisor. The student should prefer matching supervisors that have a closely related topic higher in their rank of preferences as it is the most preferred topic for the student. This fact is reflected by the definition of the rank similarity:
\begin{equation}
S_{rnk}(kw_{i},kw_{j}, \mathcal{KW}_i, \mathcal{KW}_j ) = \frac{1}{1 + | pos(kw_i,\mathcal{KW}_i) - pos(kw_j,\mathcal{KW}_{j}) | }
\end{equation}
where the function $pos$ returns the position of a keyword in a ranked list of preferences, with lower positions representing choices higher in rank. This similarity metric reflects the fact that the positions of the keyword and best matching keyword in both supervisors and students is important.
\end{itemize}

From this point on, we define the evaluation given by a student $s_i$ to a supervisor $r_j$: $V_{i,j}$. The evaluation given by $s_i$ to $r_j$ is defined as:
\begin{equation}
\label{pref-stu}
V_{i,j} = \underset{kw_{i,r} \in \mathcal{KW}_i}{\sum} w_r \times S_{rnk}(kw_{i,r},kw^*_{i,r}, \mathcal{KW}_i, \mathcal{KW}_j) \times S_{\mathcal{T}}(kw_{i,r},kw^*_{i,r}) 
\end{equation}
where, as mentioned, we define $kw^*_{i,r} \in \mathcal{KW}_j$ as the best matching topic to $kw_i$, and $w_r$ is a weight indicating the importance of matching the $r$-th most important preference for the student. This way, we take into consideration that students may prefer to be matched according to their most preferred topic rather than topics further down in their ranked list of preferences. It should be highlighted that the evaluation given by a supervisor $r_j$ to the student $s_i$, $V'_{j,i}$, can be defined in an analogous way.

\subsection{Optimization problem}
As we have mentioned in the previous section, in this article we consider both the preferences of the students and supervisors. This means that there are two objectives to be maximized and this is a multi-objective optimization problem. We describe this optimization problem first from the point of view of students and then from the point of view of the supervisors.

\subsubsection{Optimization: Student perspective}

From the point of view of the students, the optimization problem is to maximize the overall satisfaction of the students from their assigned supervisors. Next, we define the associated optimization problem:

\begin{eqnarray}
\label{eq:opt-st}
\max \;\; \frac{1}{|\mathcal{S}|} \;\; \underset{s_i \in \mathcal{S}}{\sum} \;\; \underset{r_j \in \mathcal{R}}{\sum} x_{i,j} \times V_{i,j} \nonumber \\
\mbox{subject to} \nonumber \\
\forall r_j \in \mathcal{R}, c_{j,min} \leq \underset{s_i \in \mathcal{S}}{\sum} x_{i,j} \leq c_{j,max}\\
\forall s_i \in \mathcal{S}, \underset{r_j \in \mathcal{R}}{\sum} x_{i,j} = 1 \nonumber \\
0 \leq x_{i,j} \leq 1 \nonumber
\end{eqnarray}

where $x_{i,j}$ is a binary variable that indicates if the student $s_i$ has been allocated to supervisor $r_j$ in matching $M$. The optimization function, which we aim to maximize, is defined as the mean of the valuation given by students for their assigned supervisors in matching $M$. The first constraint forces the optimization problem to find a solution where no supervisor $r_j$ is over his/her upper bound supervision quota  $c_{j,max}$, and that a minimum of $c_{j,min}$ students are allocated to $r_j$. This latter value represents situations where the department establishes a minimum supervision workload for supervisors. The next constraint forces the optimization problem to assign a student $s_i$ to just one supervisor. Finally, the last constraint defines the domain for the binary variables.

\subsubsection{Optimization: Supervisors' perspective}
As mentioned above, the other optimization problem is defined by the interests of the supervisors. In this article, we assume that the interests of the department are (i) to make supervisors more comfortable with their work by assigning a student who is willing to work in areas related to the supervisor's expertise; (ii) and to avoid unbalanced solutions where well-known supervisors have a much higher supervision load than other staff members, as this could cause friction and envy amongst coworkers. Given these assumptions, we define the optimization problem faced by the department as follows:

\begin{eqnarray}
\label{eq:opt-sup}
\max \;\; \frac{1}{(1+\sigma_{\mathcal{L}_{M}})^{\alpha}} \times \frac{1}{|\mathcal{R}|} \underset{r_j \in \mathcal{R}}{\sum} \frac{1}{|M(r_j)|} \;\; \underset{s_i \in \mathcal{S}}{\sum} \;\; x_{i,j} \times V'_{j,i} \nonumber \\
\mbox{subject to} \nonumber \\
\forall r_j \in \mathcal{R}, c_{j,min} \leq \underset{s_i \in \mathcal{S}}{\sum} x_{i,j} \leq c_{j,max}\\
\forall s_i \in \mathcal{S}, \underset{r_j \in \mathcal{R}}{\sum} x_{i,j} = 1 \nonumber \\
0 \leq x_{i,j} \leq 1 \nonumber
\end{eqnarray}

The optimization problem is similar to that defined for the students. In fact, the constraints of the problem are identical to the ones defined for the student problem.  The objective function evaluates the satisfaction of a supervisor as the average evaluation value given by the supervisor for the students allocated to him/her. Then, the overall satisfaction of supervisors is taken as the average satisfaction of all supervisors. 

Additionally, the objective function is penalized by an external factor that depends on the standard deviation of the workload levels $\sigma_{\mathcal{L}_M}$ in $M$. The greater the standard deviation, the greater becomes the penalization factor and the more is reduced the value of the objective function. As a result, given two allocations with the same value stemming from the value given by students assigned to supervisors, the optimization problem prefers solutions with a more balanced workload level. This avoids situations like the one mentioned above, where popular supervisors are highly subscribed while others have a very low workload level. The effect of this parameter can be further expanded by the coefficient $\alpha$, which should penalize allocations with a higher workload unbalance when $\alpha>1$.

\section{A Pareto optimal genetic algorithm for the  student-supervisor allocation problem}
\label{ref:GA}

Due to the ability of dealing with large search spaces and providing good solutions in a reasonable amount of time, we decide to use genetic algorithms to solve the student-supervisor allocation problem. In this section, we describe the design and implementation of the proposed genetic algorithm. In addition to this, metaheuristics tend to provide a good solution in a reasonable amount of time. Exact methods for non-linear, or even linear problems such as the one presented in Section \ref{ref:prob} are known to be costly in time for large search spaces. Thus, we select genetic algorithms as the method for solving the problem presented in this work.  

As the reader may have observed, there are two different objective functions for the problem. Therefore, in this article we opt for a Pareto optimal genetic algorithm. Pareto optimal methods allow to retrieve a variety of non-dominated solutions which can later be analyzed by a decision maker to trade-off between the different objective functions. In this case, the staff entitled with the task of allocating students to supervisors can select from a wide range of allocations to better reflect the priorities of the students and the supervisors. More specifically, due to the fact that the problem is composed by just two objective functions, we employ a schema inspired by NSGA-II \cite{nsga}, a well-known GA schema for estimating Pareto optimal solutions in multi-objective problems.

Next, we define the specific details of the proposed genetic algorithm. First, we explain how chromosome (solutions) are represented in our GA. Next, we define the main operators of the proposed GA: crossover operators, and mutation operator. Finally, we describe the selection mechanism employed and the outline of the GA.

\subsection{Chromosome representation}
For this GA, we employ a graph to represent a solution. Formally, a matching $M$ can be represented by means of a bipartite graph $G_{M}=(\mathcal{S},\mathcal{R},E)$ where $E=\{(s_i,r_j) | M(s_i) = r_j\}$ is the set of edges that determine the assignment of students to supervisors (i.e., an edge is present if a student is matched to a supervisor in M).

\begin{figure}
\centering
\includegraphics[width=0.4\linewidth]{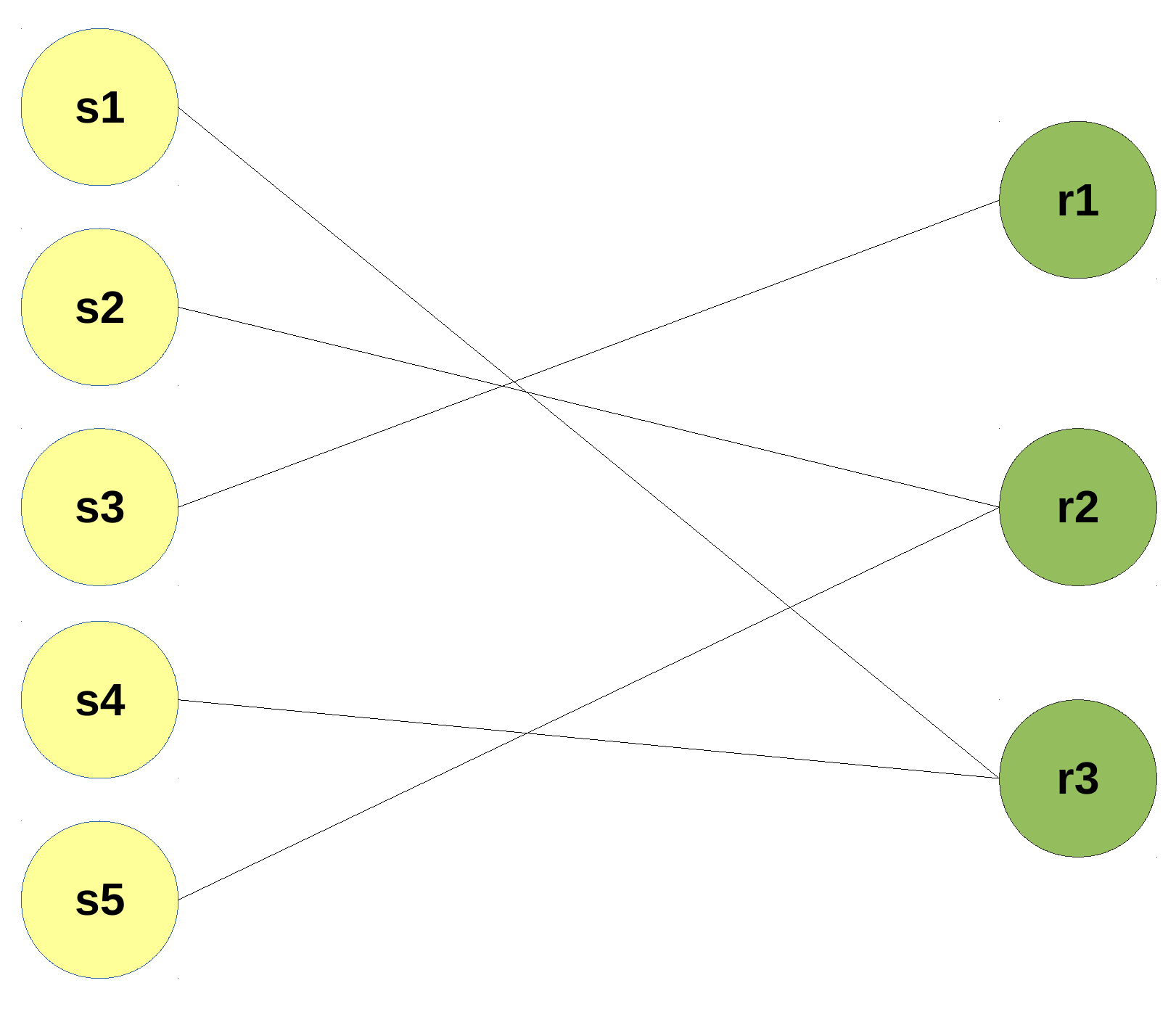}
\caption{An example of a matching of 5 students to 3 supervisors represented by a bipartite graph}
\label{img:bipartite}
\end{figure}

Figure \ref{img:bipartite} show an example of how an allocation of 5 students to 3 supervisors is represented by a bipartite graph. More specifically, in the example provided, the supervisor $r_1$ supervises student $s_3$, supervisor $r_2$ supervises students $s_2$ and $s_5$, and supervisor $r_3$ supervises students $s_1$ and $s_4$.

In the bipartite graph $G_M$ representing an allocation, we define the structure of the allocation as $st_{G_M} = (|N(r_1,G_M)|,\dots,|N(r_m,G_M)|)$ as the number of neighbors of each supervisor in the bipartite graph (i.e., the number of students that each supervisor supervises in the allocation). For instance, in the example in Figure \ref{img:bipartite}, the structure of the allocation is (1,2,2). As the reader may have guessed by now, the structure of the allocation is important as it is related to the workload level of the supervisors and, therefore, to the objective function of the supervisors.

\subsection{Mutation operator}
\label{mutation}

We introduce a mutation operator in the context of the student-supervisor allocation problem that employs two actions: \textit{swap} and \textit{transfer}. Our swap action is inspired by the mutation operator in bin packing problems \cite{Syswerda-1991}. Similarly, the transfer action is inspired on the similar intuition proposed in \cite{Falkenauer-1992} for bin packing problems.

The mutation operator is applied over a single parent and it generates a single child. For this problem we have designed a special mutation operator that applies a series of operations on a parent allocation: swapping of students between supervisors, and transferring of a student from one supervisor to another. The former operation does not change the structure of the allocation (i.e., the workload of any supervisor), while the latter does by reducing the workload of a supervisor by one and increasing the load of another supervisor by one.

The extent to which a parent changes by a single mutation operation is defined by the mutation ratio $p_{mt}$, that represents the probability of mutating an edge in a graph $G_M$. The type of operation that is applied over an edge that is to be mutated is controlled by $p_{sw}$, which controls the probability of applying a swapping operation between two supervisors. An outline of the mutation operator can be found in Algorithm \ref{alg:mut}. The operator iterates over edges in the bipartite graph and attempts to perform an operation over an edge in the graph with a probability of $p_{mt}$. In case that a transfer operation is possible for the edge (i.e., the supervisor has more than the minimum quota established by the department), it selects a random supervisor that can take students (i.e., under-subscribed) and performs the operation with a probability of $1-p_{sw}$ (lines 1 to 9). Otherwise, the operation selected is a swap between supervisors (lines 10 to 15), which is always possible in  feasible solution as the structure of the solution does not change.
        
\begin{algorithm}
\small
  \KwIn{$G_M = (\mathcal{S},\mathcal{R},E):$ A bipartite graph representing a feasible allocation}
  \KwOut{$G_{M'} = (\mathcal{S},\mathcal{R},E'):$ A new bipartite graph representing a feasible allocation}
  $G_{M'} = G_M$\;
  $\mathcal{U} = \mbox{under-subscribed}(G_{M'})$\;
  \For{ $(s_i,r_j) \in E'$ }{
    \If{ $random() \leq p_{mt}$ }{
    	\eIf{ $random() > p_{sw} \; \wedge \; N(r_j,G_{M'}) > c_{j,min} \wedge \; \mathcal{U} \neq \emptyset   $}
        {
        	\textit{/*Transfer operation*/}\;
        	$r_{q} = random\_choice(\mathcal{U})$\;
            $E'= (E' \cup \{(s_i,r_q)\} ) - \{(s_i,r_j)\}$\;
            $\mathcal{U} = \mbox{under-subscribed}(G_{M'})$\;
        }  
        {
        	\textit{/*Swap operation*/}\;
        	$r_q = random\_choice(\mathcal{R})$\;
            $s_p = random\_choice(N(r_q,G_{M'}))$\;
            $E'=(E'\cup \{ (s_i,r_q), (s_p,r_j) \}) - \{ (s_i,r_j), (s_p,r_q)  \}$\;
        }
    }
  }
\caption{Outline of the mutation operator}
\label{alg:mut}
\end{algorithm}

\subsection{Crossover operators}
\label{sec:xover}

As the structure of the allocation is important for the objective function of the supervisors, we have devised two crossover operators for matching problems that preserve the allocation structure of the parents as much as possible. More specifically, these two crossover operators take two parents as input and produce a child as a result. Both operators preserve the original allocation structure of one of the parents; however, the second approach may end up adding new genetic material not present in any of the parents whereas the first one does not. As a result, the latter crossover operation may induce in additional exploration as new assignments of students to supervisors. We call the first genetic operator as the \textit{Hopcroft-Karp} genetic operator as it is based on the popular algorithm to find maximum cardinality matchings in bipartite graphs \cite{hopcroft}, while the second receives the name of \textit{greedy structural preservation} crossover due to its greedy nature for selecting genetic material from parents. Next, we define both in detail.

\subsubsection{Hopcroft-Karp genetic operator}

As we mentioned, this crossover operator generates a new child from two parents. In order to do so, the new solution inherits the allocation structure of one of the parents, and it exclusively employs genetic material from the two parents to generate the child.

\begin{figure}
\centering
    \begin{subfigure}[t]{0.49\textwidth}
        \raisebox{-\height}{\includegraphics[width=\textwidth]{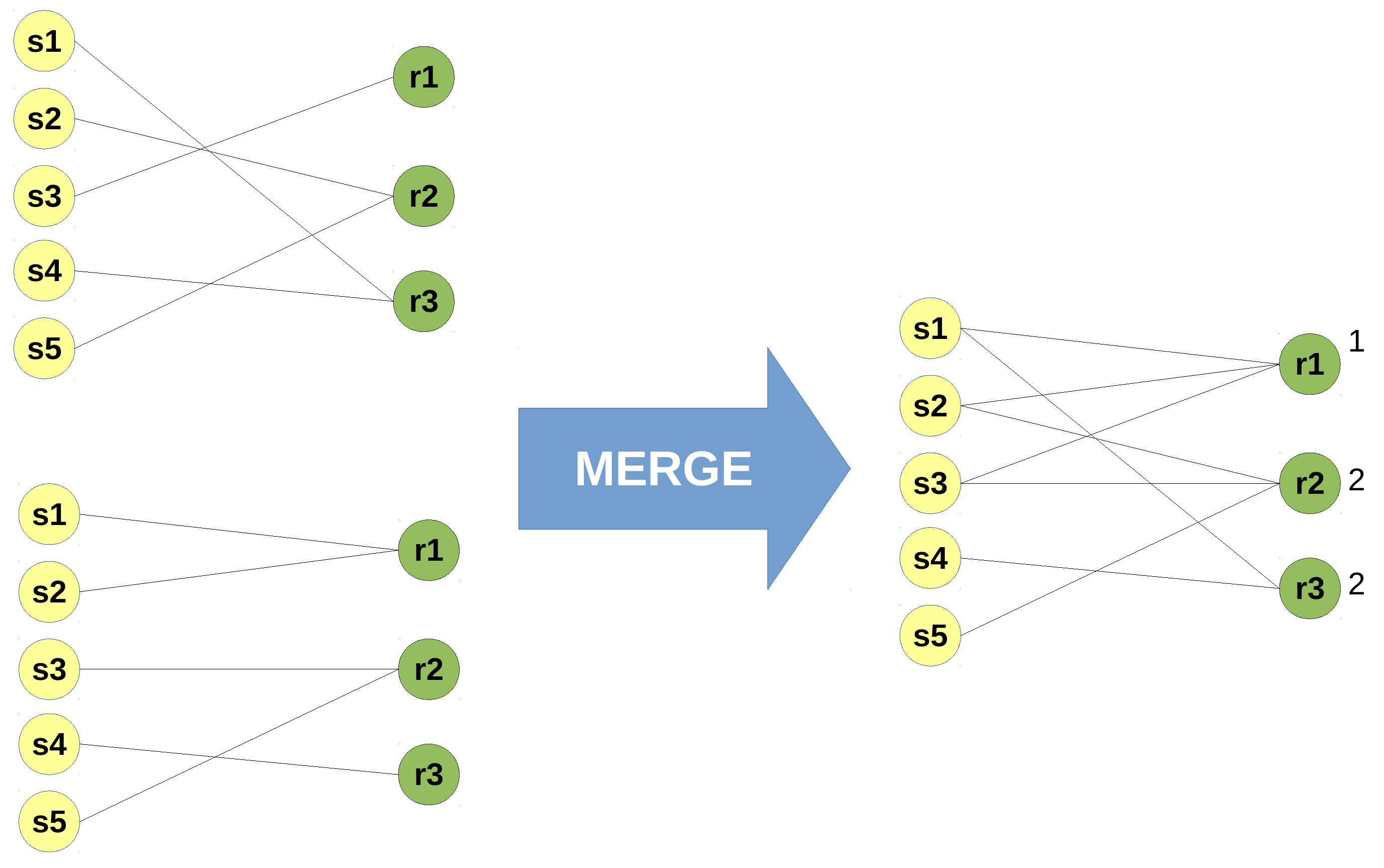}}
        \caption{}
    \end{subfigure}
    \hfill
    \begin{subfigure}[t]{0.49\textwidth}
        \raisebox{-\height}{\includegraphics[width=\textwidth]{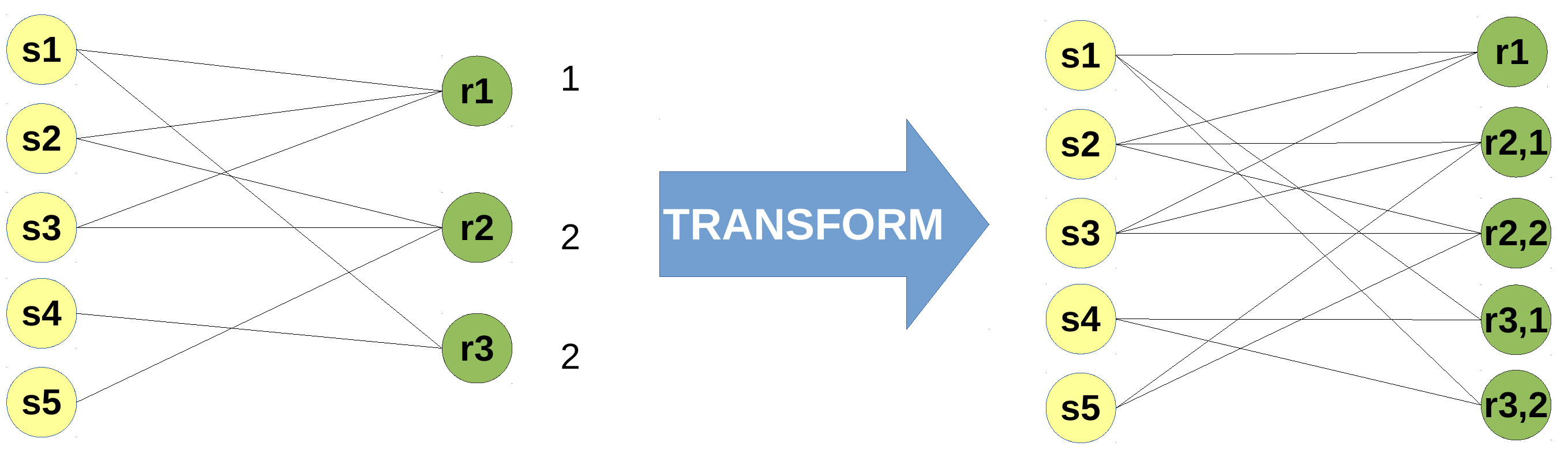}}
        \caption{}
    \end{subfigure}
    \begin{subfigure}[t]{0.49\textwidth}
        \raisebox{-\height}{\includegraphics[width=\textwidth]{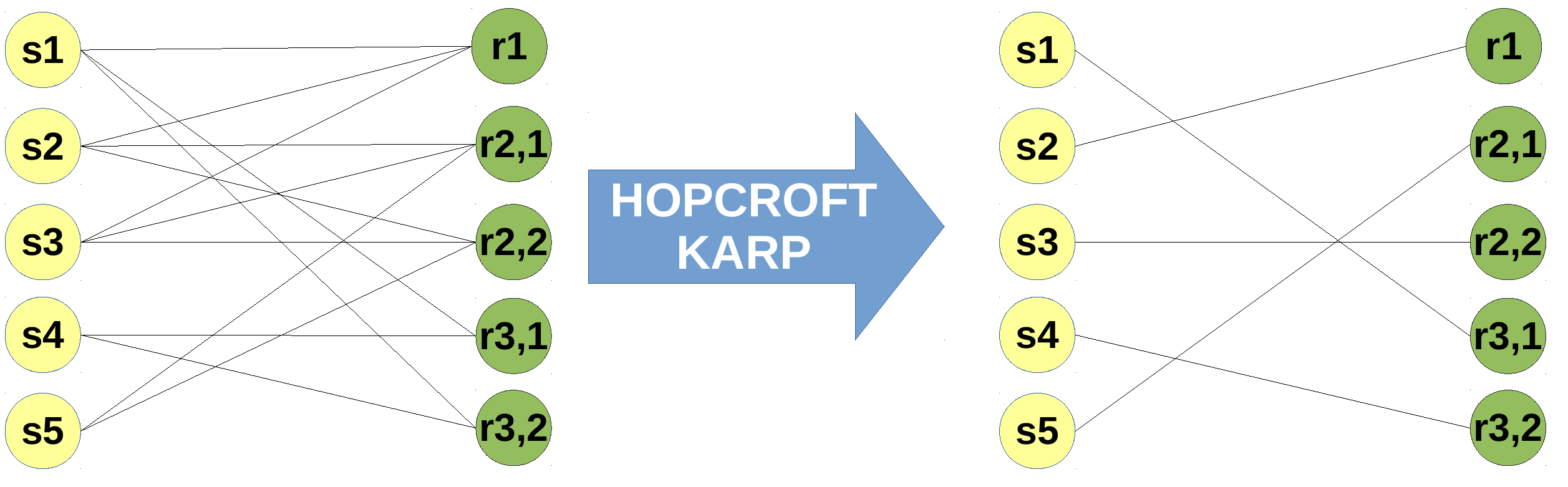}}
        \caption{}
    \end{subfigure}
    \hfill
    \begin{subfigure}[t]{0.49\textwidth}
        \raisebox{-\height}{\includegraphics[width=\textwidth]{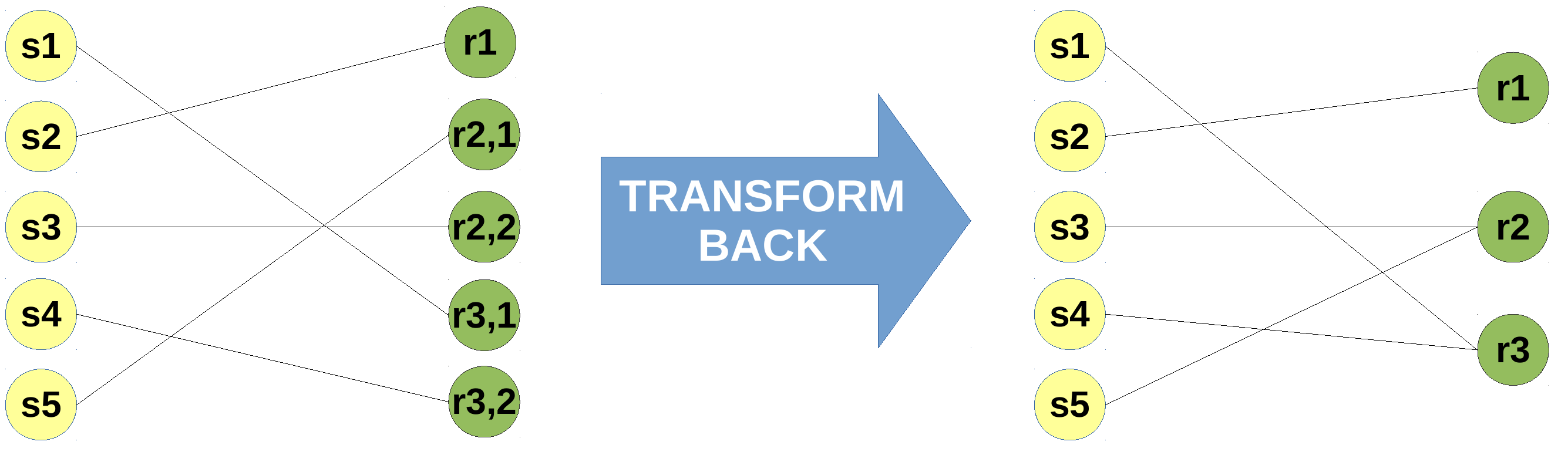}}
    \caption{} 
    \end{subfigure}
    \caption{Steps of the \textit{Hopcroft-Karp} crossover operator (in separate boxes). The order of the steps is read left to right.}
    \label{hop}
\end{figure}

The outline of this genetic operator is as follows. An example of the application of this genetic operator to two parents can be found in Figure \ref{hop}, while the formalization of the operator can be found in Algorithm \ref{alg:hop}: 
\begin{itemize}


	\item \textbf{Merging parents:} This step can be found in Figure \ref{hop} (a) and lines 1-10 of Algorithm \ref{alg:hop}. First of all, we generate a new graph as a result of merging both graphs by keeping the set of students and supervisors. The selection of this structure is proportional to its impact in the objective function of the supervisors. In addition to this, the structure of one of the two parents is inherited as a goal for the new bipartite graph. In the example in Figure \ref{hop}, the structure of the first parent is chosen.

    \item \textbf{Transforming graph:} The description of this step can be found in lines 11-14 of Algorithm \ref{alg:hop} and an example can be found in Figure \ref{hop} (b). The merged graph is transformed into a new bipartite graph whose set of supervisors contains a copy of each original supervisor for each student that he/she should supervise according to the inherited structure. For instance, in the example in Figure \ref{hop} (b), there are two copies of the original supervisor $r_2$ ($r_{2,1}$ and $r_{2,2}$) because two students should be assigned to the second supervisor in the new allocation. Similarly, the same happens for supervisor $r_{3}$ ($r_{3,1}$, $r_{3,2}$).
    \item \textbf{Hopcroft-Karp:} Lines 15-16 of Algorithm \ref{alg:hop} and Figure \ref{hop} (c) represent this step. The Hopcroft-Karp algorithm \cite{hopcroft} is applied on the transformed graph to find a maximum cardinality matching. As the merged graph contains at least a perfect matching (i.e., one of the two original parents), then the maximum cardinality matching is a perfect matching.

    \item \textbf{Transforming back:} Finally, the perfect matching is transformed back to the original supervisor set by merging those nodes that represent the same supervisor. As a result of this process, a new child is generated that inherits the structure of one of the two parents and it introduces no new genetic material. This can be found in lines 17-19 of Algorithm \ref{alg:hop} and Figure \ref{hop} (d). Note that the structure of the allocation is the same with one of the parents; however, it does not mean that the allocation is exactly the same with the chosen parent's allocation. As seen from the given example, the resulted allocation is different than the parent allocations (e.g. s3 is assigned to r1 in the chosen parent allocation while it is assigned to r2 in the child allocation).
\end{itemize}     

The theoretical complexity of this crossover operator is determined by the complexity of each of its individual steps. In order to merge both parents, a new set of edges must be created which consists of all the edges in both parents. As the number of edges in each parent is exactly  $|\mathcal{S}|$ then the cost of this step is $\mathcal{O}(|\mathcal{S}|)$. Transforming the graph requires to create a new set of supervisors that has exactly as many supervisors as students ($\mathcal{O}(|\mathcal{S}|)$) and creating a new set of edges that is at most $\mathcal{O}(|\mathcal{S}|)$. The most expensive step is applying the Hopcroft-Karp algorithm which has a complexity of $\mathcal{O}(|\mathcal{S}|\;\sqrt[]{ |\mathcal{S}| })$ in the worst case. However, some recent studies show that in the average case the Hopcroft-Karp algorithm has a complexity of $\Theta(|\mathcal{S}| log |\mathcal{S}| )$ for random sparse bipartite graphs \cite{bast2006matching}. The bipartite graphs generated by the merge operation will result in graphs where students have at most two neighbors (i.e., the student has a different supervisor in both parents). Therefore, we expect that in practice the cost of this step will be closer to the $\Theta(|\mathcal{S}| log |\mathcal{S}|)$  complexity. The final step requires iterating over resulting edges in the perfect matching which is exactly $\mathcal{O}(|\mathcal{S}|)$. Therefore, the complexity of this operator is $\mathcal{O}(|\mathcal{S}|\sqrt{|\mathcal{S}|})$ in the worst case and we expect it to be $\Theta(|\mathcal{S}| log |\mathcal{S}|)$ in the average case. In both cases, the complexity is quasi-linear. 

\begin{algorithm}[t]
\small
  \KwIn{$G_{M_1} = (\mathcal{S},\mathcal{R},E_1):$ A bipartite graph representing a feasible allocation; $G_{M_2} = (\mathcal{S},\mathcal{R},E_2):$ A bipartite graph representing a feasible allocation;}
  \KwOut{$G_{M'} = (\mathcal{S},\mathcal{R},E'):$ A new bipartite graph representing a feasible allocation}
\textit{/* Merge graphs */} \;
$G = (\mathcal{S},\mathcal{R}, E=E_1 \cup E_2)$\;
\textit{/* Inherit one of the structures */}\;
$p_1 = \frac{1}{(1+\sigma_{M_1})^\alpha}$\;
$p_2 = \frac{1}{(1+\sigma_{M_2})^\alpha}$\;
\eIf{ $random() \leq \frac{p_1}{p_1 + p_2}$ }{
	$st_{G_{M'}} =  \{|N(r_1,G_{M_1})|,\dots,|N(r_m,G_{M_1})|\}$\;
}
{
$st_{G_{M'}} =  \{|N(r_1,G_{M_2})|,\dots,|N(r_m,G_{M_2})|\}$\;
}

\textit{/* Transform graph */} \;
$\mathcal{R}_{tr} = \{ r_{j,l} \;|\; r_{j} \in \mathcal{R} \; \wedge \;  l \leq st_{G_{M'}}(r_j)  \}$\;
$E_{tr} = \{ (s_i,r_{j,l}) \; | \; ((s_i,r_j) \in E_1 \cup E_2) \; \wedge \; r_{j,l} \in \mathcal{R}_{tr}  \}$ \;
$G_{tr} = ( \mathcal{S}, \mathcal{R}_{tr} , E_{tr} )$ \;
\textit{/* Apply Hopcroft-Karp algorithm */} \;
$E_{hp} = hopcroft\_karp(G_{tr})$\;
\textit{/* Transform back to original representation */}\;
$ E' = \{ (s_i,r_j) \; | \; \exists (s_i,r_{j,l}) \in E_{hp} \} $\;
$G' = (\mathcal{S}, \mathcal{R}, E' )$
\caption{The Hopcroft-Karp crossover operator}
\label{alg:hop}
\end{algorithm}

\subsubsection{Greedy structural preservation genetic operator}

The greedy approach preserves the structure of one of the two parents, which is randomly inherited based on the impact of the structure on the fitness of the supervisors. Differently to the \textit{Hopcroft-Karp} crossover operator, this crossover operator may introduce new genes that are not present in any of the two parents. Nevertheless, the operator aims to keep original genetic material as much as possible. As a trade-off, the computational complexity of this operator is lower than that of the \textit{Hopcroft-Karp} as it takes a greedy approach. The general idea behind this method is locking edges that will be part of the resulting matching, and removing those that are to be discarded.

\begin{figure}
\centering
    \begin{subfigure}[t]{0.49\textwidth}
        \raisebox{-\height}{\includegraphics[width=\textwidth]{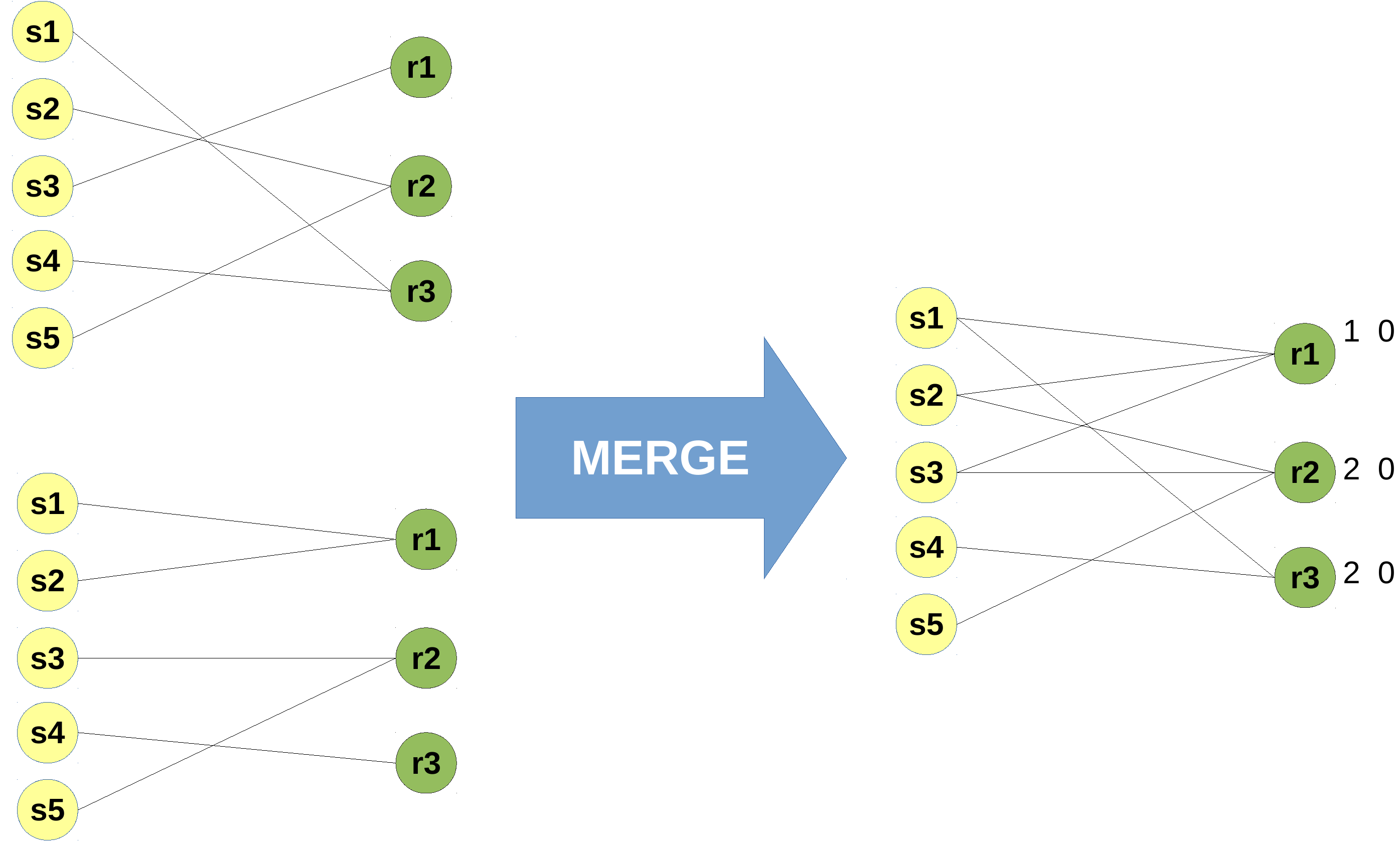}}
        \caption{}
    \end{subfigure}
    \hfill
    \begin{subfigure}[t]{0.49\textwidth}
        \raisebox{-\height}{\includegraphics[width=\textwidth]{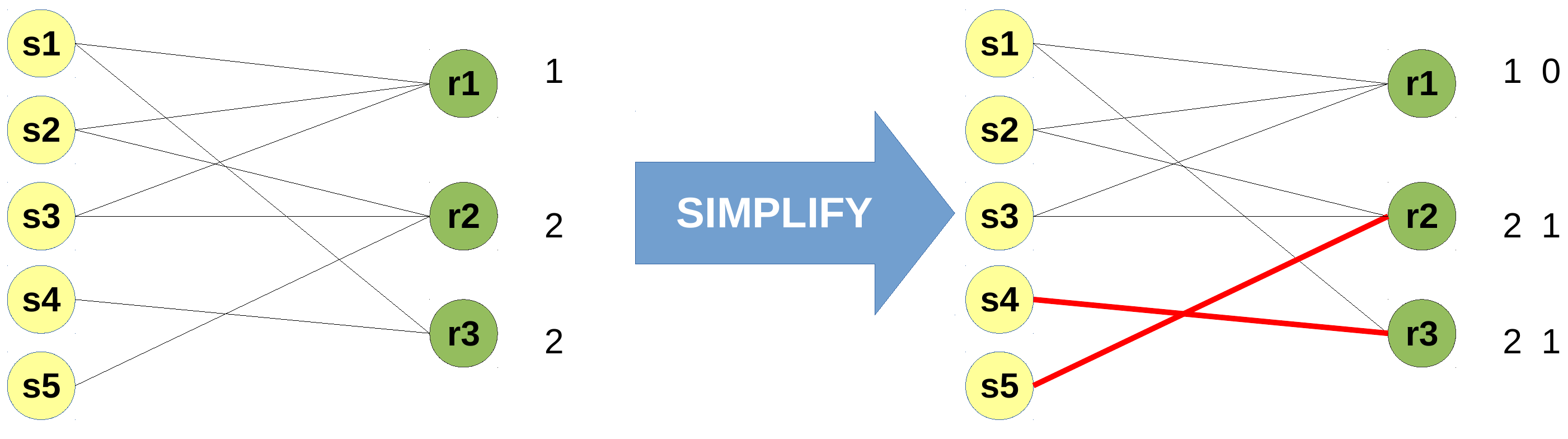}}
        \caption{}
    \end{subfigure}
    \begin{subfigure}[t]{0.49\textwidth}
        \raisebox{-\height}{\includegraphics[width=\textwidth]{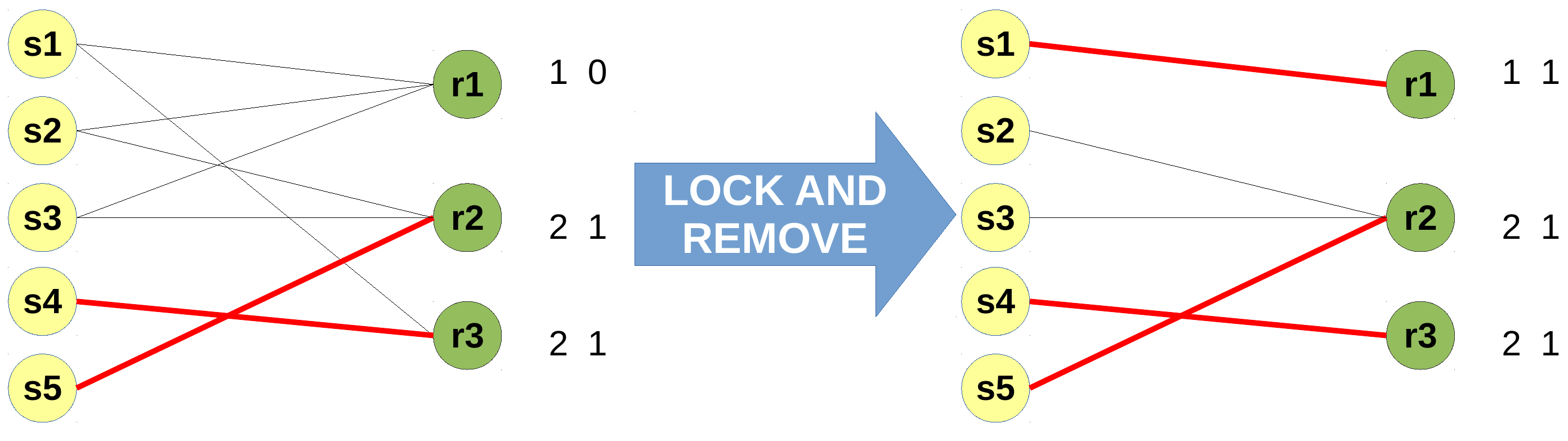}}
        \caption{}
    \end{subfigure}
    \hfill
    \begin{subfigure}[t]{0.49\textwidth}
        \raisebox{-\height}{\includegraphics[width=\textwidth]{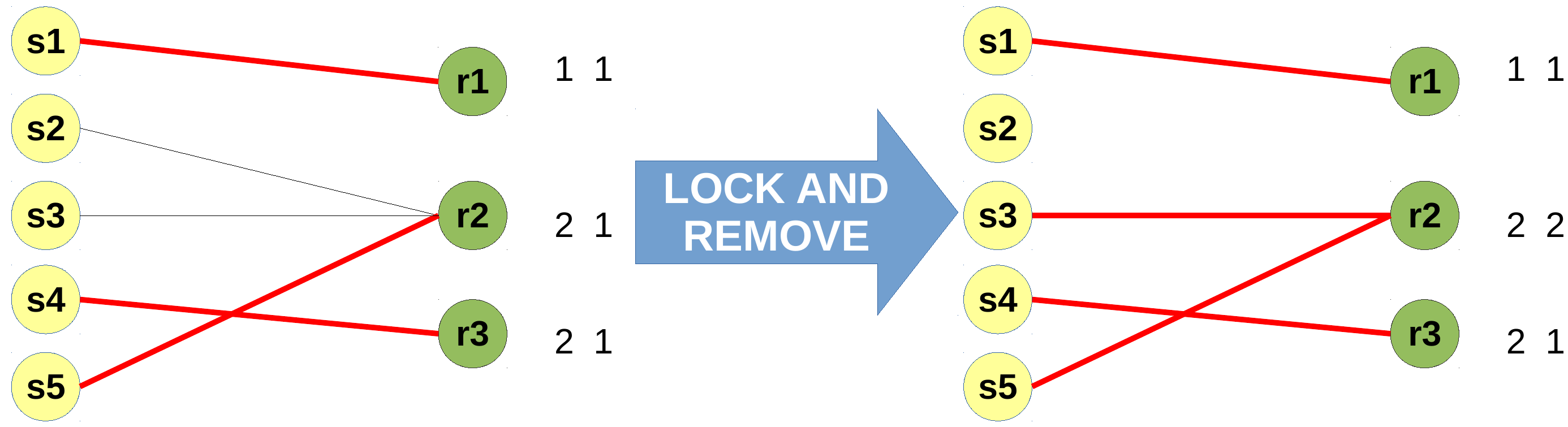}}
    \caption{} 
    \end{subfigure}
  \begin{subfigure}[t]{0.49\textwidth}
        \raisebox{-\height}{\includegraphics[width=\textwidth]{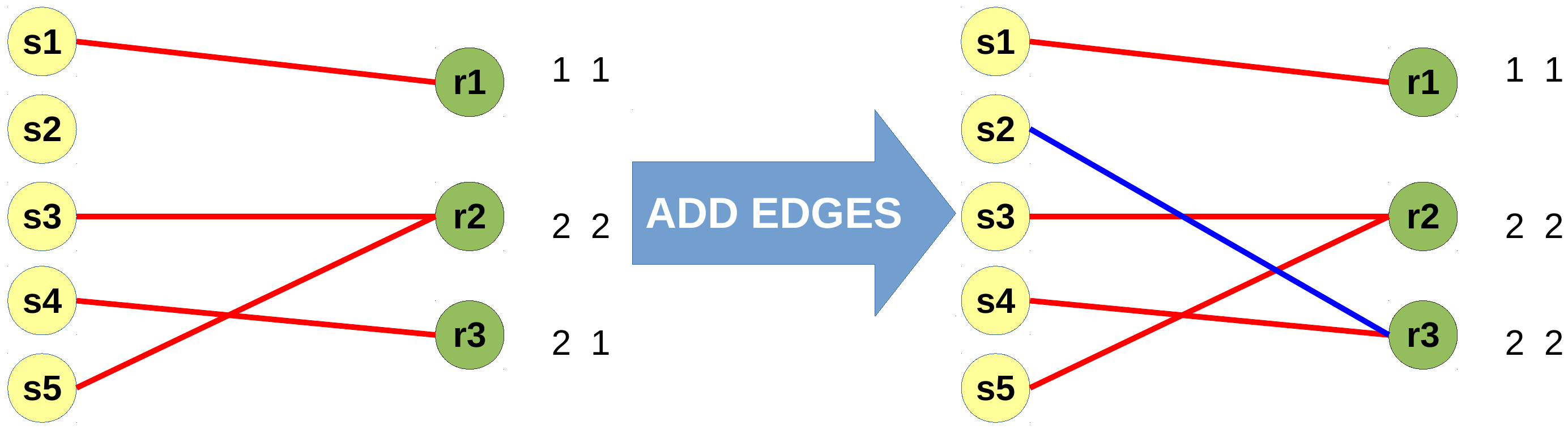}}
        \caption{}
    \end{subfigure}
    \hfill
        \caption{Steps of the \textit{greedy structural preservation} crossover operator (in separate boxes). The order of the steps is read left to right.}
    \label{greedy}
\end{figure}

Next, we describe the outline of this genetic operator in more detail. Figure \ref{greedy} shows how the operator is applied over a particular example, while Algorithm \ref{alg:greedy} depicts the specific details of the operator in more detail. The general steps of the crossover operator are:
\begin{itemize}

	\item \textbf{Merging parents:} This step can be found in the top left box of Figure \ref{greedy} (a) and lines 1-11 of Algorithm \ref{alg:greedy}. It is equivalent to the merging steps in \textit{Hopcroft-Karp} crossover operator. In addition to inheriting the structure of one of the two solutions, the method initializes a counter for each supervisor that contains the number of edges that have been locked for the final allocation so far in the process. In Figure \ref{greedy} (a), the allocation structure inherited is that from the first parent.
  
    \item \textbf{Simplify:} The details of this step can be found in Algorithm \ref{alg:greedy} from lines 12 to 20, and an example applied over a real graph is observable in Figure \ref{greedy} (b). The merged bipartite graph is simplified. The simplification process locks those edges that have a student with a single possible supervisor. For instance, in Figure \ref{greedy} (b), this corresponds to edges starting from students $s_4$ and $s_5$. The locked edges will be part of the final allocation and counters are updated for supervisors whose edges have been locked ($r_2$ and $r_3$). In case that one of the supervisors reaches the desired workload level, all other unlocked edges involving that supervisor will be removed from the merged graph. This step is repeated until the graph cannot be further simplified by this method.
    \item \textbf{Locking and removing edges:} This step corresponds to lines 21 to 30 in Algorithm \ref{alg:greedy}, and Figure \ref{greedy} (c).  An unlocked edge is randomly chosen from the current graph (e.g., edge $(s_1,r_1)$ in Figure \ref{greedy} (c)) and it is locked to be part of the final allocation. As the edge has been locked, the number of locked edges for the supervisor is increased. Any other edges incident in the selected student are removed from the merged graph (e.g., edge $(s_1, r_3)$ in Figure \ref{greedy} (c)). In case that the supervisor has reached the desired workload level, unlocked edges incident in the supervisor are removed from the graph (e.g., edges $(s_2, r_1)$ and $(s_3, r_1)$ in Figure \ref{greedy} (c)). This step is repeated while there are no more unlocked edges.
    \item \textbf{Adding edges:} This last step is represented in lines 31 to 39 of Algorithm \ref{alg:greedy} and Figure \ref{greedy} (e). Once there are no more unlocked edges, the operator checks if there are any unmatched students and supervisors. If there are unmatched vertex, then the operator randomly assigns students to supervisors while following the desired workload level in the allocation, and considering the number of locked edges for each supervisor. The process of adding edges is repeated until there are no more unmatched students.
\end{itemize}

The process described above ends up with a feasible allocation which has inherited the structure of one of the two parents. The complexity of the operator is straightforward. As discussed, the merging process has a complexity proportional to $\mathcal{O}(|\mathcal{S}|)$. In the worst case, the simplify step will be applied as many times as students in the problem (i.e., merging the same parents) which gives a worst case cost of $\mathcal{O}(|\mathcal{S}|)$. The lock and remove step will be applied as any times as edges in the merged graph, which will be $\mathcal{O}(|\mathcal{S}|)$ in the worst case. Then, the final step adds one edge per remaining unassigned student. This last step will never be more costly than $\mathcal{O}(|\mathcal{S}|)$. Therefore, the cost of this operator is linear with the number of students in the problem.

\begin{algorithm}
\small
  \KwIn{$G_{M_1} = (\mathcal{S},\mathcal{R},E_1):$ A bipartite graph representing a feasible allocation; $G_{M_2} = (\mathcal{S},\mathcal{R},E_2):$ A bipartite graph representing a feasible allocation;}
  \KwOut{$G_{M'} = (\mathcal{S},\mathcal{R},E'):$ A new bipartite graph representing a feasible allocation}
\textit{/* Merge graphs */} \;
$G' = ( \mathcal{S}' = \mathcal{S}, \mathcal{R}' = \mathcal{R}, E'=E_1 \cup E_2)$\;
\textit{/* Inherit one of the structures */}\;
$p_1 = \frac{1}{(1+\sigma_{M_1})^\alpha}$\;
$p_2 = \frac{1}{(1+\sigma_{M_2})^\alpha}$\;
\eIf{ $random() \leq \frac{p_1}{p_1 + p_2}$ }{
	$st_{G_{M'}} =  \{|N(r_1,G_{M_1})|,\dots,|N(r_m,G_{M_1})|\}$\;
}
{
$st_{G_{M'}} =  \{|N(r_1,G_{M_2})|,\dots,|N(r_m,G_{M_2})|\}$\;
}
$\mathcal{L} = \emptyset$ \textit{/* Initializing locked edges */ }\;
\textit{/* Simplify graph */} \;
\ForEach{ $ \{ (s_i,r_j)  \;\;|\;\; |N(s_i,G_{M'})| = 1 \}$  }
{
 $\mathcal{L} = \mathcal{L} \cup \{(s_i,r_j)\}$\;
 $l_j = l_j + 1$ \textit{/* Update locked edges counter for supervisor $j$ */} \;
 \textit{/* If supervisor has desired workload level, then remove non-locked edges */}\;
 \If{ $l_j = st_{G_{M'}}(j)$ }{
 	$E'= E' - \{ (s_u, r_j) \;\; | \;\; (s_u,r_j) \notin \mathcal{L} \}$
 }
}
\textit{ /* Locking and removing edges */ }\;
\While{ $E' - \mathcal{L} \neq \emptyset$ }{
	$(s_i,r_j) = random\_choice( E' - \mathcal{L} )$\;
    $E' = E' - \{ (s_i,r_l) \;\; | r_l \neq r_j \}$ \textit{ /* Remove other edges incident in the student */} \;
    $\mathcal{L} = \mathcal{L} \cup \{ (s_i,r_j) \}$\;
    $l_j = l_j + 1$\;
    \If{ $l_j = st_{G_{M'}(j)}$ }{ 
    	$E'= E' - \{ (s_u, r_j) \;\; | \;\; (s_u,r_j) \notin \mathcal{L} \}$ 
    }
}
\textit{ /* Adding edges to complete graph */ }\;
$ \mathcal{S}_{re} = \{ s_i \;\; | \;\; |N(s_i,G_{M'})| = 0 \}  $\;
$ \mathcal{R}_{re} = \{ r_j \;\; | \;\; l_j \neq st_{ G_{M'}(r_j) } \}  $\;
\While{ $\mathcal{S}_{re} \neq \emptyset$ }{
	$s_i = random\_choice( \mathcal{S}_{re} )$\;
    $r_j = random\_choice( \mathcal{R}_{re} )$\;
    $\mathcal{L} = \mathcal{L} \cup \{ (s_i,r_j) \}$\;
    $update( \mathcal{S}_{re}, \mathcal{R}_{re} )$\;
}
\caption{The \textit{greedy structural preservation} crossover operator}
\label{alg:greedy}
\end{algorithm}

\subsection{Selection mechanism}
The selection mechanism in this GA is employed to determine the parents that will take part in the crossover operation. More specifically, we run random tournaments \cite{goldberg91} between solutions in the population until we have selected a number of pairs that is equal to half of the current population.

As this is a multi-objective optimization problem, the comparison carried out in the tournament is determined by the solution that has a lower nondominated rank or the one that has a higher crowding distance in case of both solutions having the same nondominated rank. The nondominated rank of a solution is determined when calculating the different Pareto frontiers in the population, and it is related to the number of solutions that dominate the specific solution. On the other hand, the crowding distance makes sure that the solutions are well-spread on the Pareto frontier. The details of these metrics can be found in \cite{nsga}.

\subsection{Evolution schema}
As mentioned, the outline of the genetic algorithm is inspired by NSGA-II \cite{nsga}. The details of the GA can be found in Algorithm \ref{alg:ga}. The genetic algorithm initializes a population of $pop_{max}$ random feasible solutions (line 1). Then, the main loop of the genetic algorithm runs for a fixed number of iterations (lines 4-20), determined by the parameter $it_{max}$. 

In the main loop, the genetic algorithm calculates the successive Pareto optimal frontiers in the current population ($\mathcal{P}$, line 6): calculating the first Pareto optimal frontier, removing the Pareto frontier from the set and calculating a new one following this process until no more frontiers can be calculated. The genetic algorithm limits the number of solutions in the population by filling the new population ($\mathcal{P}_{new}$) with solutions from the first to the latest Pareto optimal frontiers (lines 8-15). Then, each solution in the resulting population is mutated, and the crossover operator is applied over solutions selected by tournament selection (lines 16-19). Lines 21 and 22 calculate the resulting frontiers after applying genetic operators in the last iteration. 

\begin{algorithm}
\small
$\mathcal{P} = \mathcal{P}_{new} = initialize( pop_{max} )$\;
$ it = 0 $\;
$\mathcal{P}_{off} = \emptyset$\;
\While{ $it < it_{max}$ }{
	$ \mathcal{P} = \mathcal{P}_{new} \cup \mathcal{P}_{off} $\;
	$\mathcal{F} = calculate\_frontiers(\mathcal{P})$\;
    $\mathcal{P}_{new} = \emptyset $\;
    \ForEach{ $f \in \mathcal{F}$}{
    	\eIf{ $|\mathcal{P}_{new}| + |f| \leq pop_{max}$  }{
        	$\mathcal{P}_{new} = \mathcal{P}_{new} \cup f$\;
        }{
        	$\mathcal{P}_{new} = \mathcal{P}_{new} \cup  select( pop_{max} - |\mathcal{P}_{new}|, f )$\;
            break\;
        }
    }
    $P_{mut} = mutation(\mathcal{P})$\;
    $\mathcal{P}' = tournament\_selection(\mathcal{P})$\;
    $\mathcal{P}_{cr} = crossover(\mathcal{P}')$\;
    $\mathcal{P}_{off} = \mathcal{P}_{mut} \cup \mathcal{P}_{cr}  $\;
    
}
$\mathcal{P} = \mathcal{P}_{new} \cup \mathcal{P}_{off}$\;
$\mathcal{F} = calculate\_frontiers(\mathcal{P})$\;
\caption{The proposed Pareto optimal genetic algorithm}
\label{alg:ga}
\end{algorithm}

\section{Experiments}
\label{ref:experiments}

In order to validate the performance of the proposed genetic algorithm, we carry out a series of practical experiments. These experiments aim to study the impact of the different elements of the genetic algorithm, as well as the overall performance of the genetic proposal. First, we provide a brief analysis of the real data collected from the student-supervisor allocation process at Coventry University, as this data is employed to create real allocation problems that will be employed to validate the performance of the genetic proposal. Then, we empirically analyze the impact of the mutation operator on the performance of the GA by studying the appropriate degree of mutation rate and the importance given to exploring the structure of the allocation rather than the allocation itself. After that, we analyze the empirical complexity of the \textit{Hopcroft-Karp} and the \textit{greedy structural preservation} crossover operator, and we compare their optimization performance with classic crossover operators. Finally, we compare the performance of the proposed genetic algorithm with that of global optimal optimization methods to assess the quality of the solutions found by the GA.

\subsection{Dataset}

In order to test the genetic algorithm in a realistic setting, we collected real data from undergraduate students and staff members that participate in the  undergraduate dissertation module for computing related degrees at Coventry University. The preferences of students and staff members were elicited by allowing individuals to specify, in order, their $k=5$ most preferred topics in the 2012 ACM Computing Classification System\footnote{\url{https://www.acm.org/publications/class-2012}}. This taxonomy provides a tree-like and hierarchical classification of areas in computing, as needed by our fitness functions, and it is a well-known system employed to categorize research papers in computing. 

A total of of 195 students' preferences and 33 supervisors' preferences were collected. This dataset\footnote{The dataset is available at \url{http://sanchez-anguix.com/index.php/research/} } contains real preferences of students on computing areas for their undergraduate dissertations, as well as the preferences of staff members on research areas where they would like to supervise students on. 

\begin{figure}
\centering
	\includegraphics[width=0.495\linewidth]{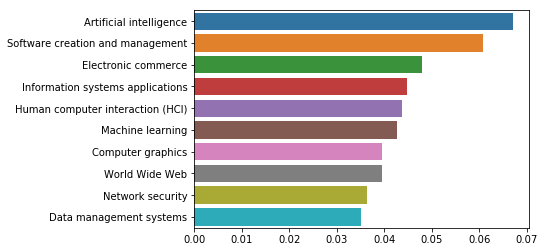}
    \includegraphics[width=0.495\linewidth]{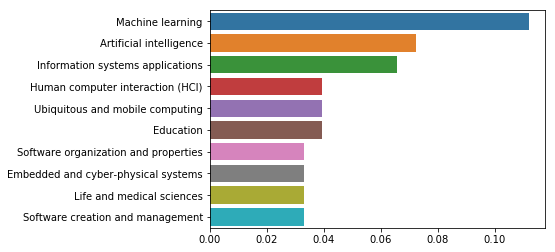}
    \caption{Distribution of the top 10 third level topics in the 2012 ACM Computing Classification System for the topics selected by students (left) and supervisors (right)}
    \label{img:hist_kw}
\end{figure}

By analyzing the preferences of both students and supervisors, one can observe that there are some differences. For instance, Figure \ref{img:hist_kw} analyzes the distribution of the top 10 topics selected by students and supervisors when focusing on the third level of the path defined by the topics selected by both populations. As one can observe, some topics that are popular amongst students like \textit{Software creation and management} are not as popular for supervisors, while some popular topics amongst students like \textit{Electronic commerce} are not even present in the top 10 third level topics for supervisors. Therefore, there is conflict between the students' and supervisors' preferences with respect to dissertation areas.

\begin{figure}
\centering
	\includegraphics[width=0.43\linewidth]{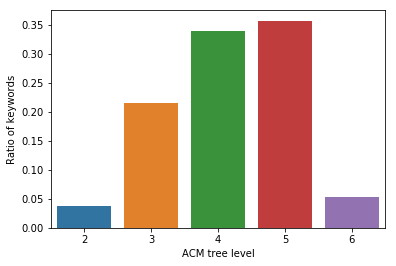}
    \includegraphics[width=0.43\linewidth]{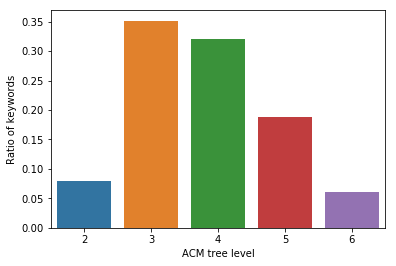}
    \caption{Level of the topics selected by students (left) and supervisors (right) from the 2012 ACM Computing Classification System}
    \label{img:hist_levels}
\end{figure}

In Figure \ref{img:hist_levels}, we analyze the level or depth of the topics provided by supervisors and students. As we can see, there are divergences with respect to the specificity of topics. The reader can observe that while supervisors were more generic with their provided topics, students were more prone to provide a fine-grained topic for their dissertations. Therefore, we can conclude again that the optimization problem is complex due to the diversity of preferences. 

It should be highlighted that the preferences contained in this dataset were employed to generate the student and supervisor profiles in the subsequent experiments.

\subsection{Optimizing the mutation operator}

As a first step to analyze the performance of the proposed genetic algorithm, we studied the impact of the mutation operator and its parameters on the problem. To be specific, we studied what the impact of the parameters $p_{sw}$ and $p_{mt}$ is on the general performance of the GA. For this matter, we created a experiment as follows:
\begin{itemize}
	\item We created 5 problems consisting of 150 students and 30 supervisors from the collected dataset. The minimum workload of supervisors $c_{j,min}$ was set to 1 student (i.e., a supervisor will advise at least one student) and the upper bound supervision quota $c_{j,max}$ of each supervisor was generated from a uniform distribution $U(4,10)$, guaranteeing that the sum of all the supervisors upper bounds exceeded in 20\% the total number of students (i.e., 180 students of capacity). 
    \item We set $\alpha=2$ to highly penalize solutions with a high standard deviation for the workload level of supervisors.
    \item The weights of topics' ranks in $V_{i,j}$ were set to (0.561,0.258,0.129,0.064,0.032) respectively, following an exponential decreasing function. This way, we take into consideration the fact that the disappointment of being matched on the second topic over the first topic is not linearly related to the difference of being matched on the last topic over the second last topic, as it was suggested by \cite{harper05}. 
    \item The crossover operation was deactivated to isolate the effect of the mutation operation on the performance of the genetic algorithm.
    \item The initial population size was set to 128 solution, and the initial population was shared amongst different runs of the same case in order to compare results on a fair basis.
    \item The maximum number of iterations $it_{max}$ was set to 250 iterations.
    \item The values tested for $p_{mt}$ ranged from 0.05 to 0.5 with increments of 0.05. On the other hand, the values tested for $p_{sw}$ ranged from 0.1 to 0.9 with increments of 0.1.
\end{itemize}

    \begin{figure}
    	\centering
    	\includegraphics[width=0.8\linewidth]{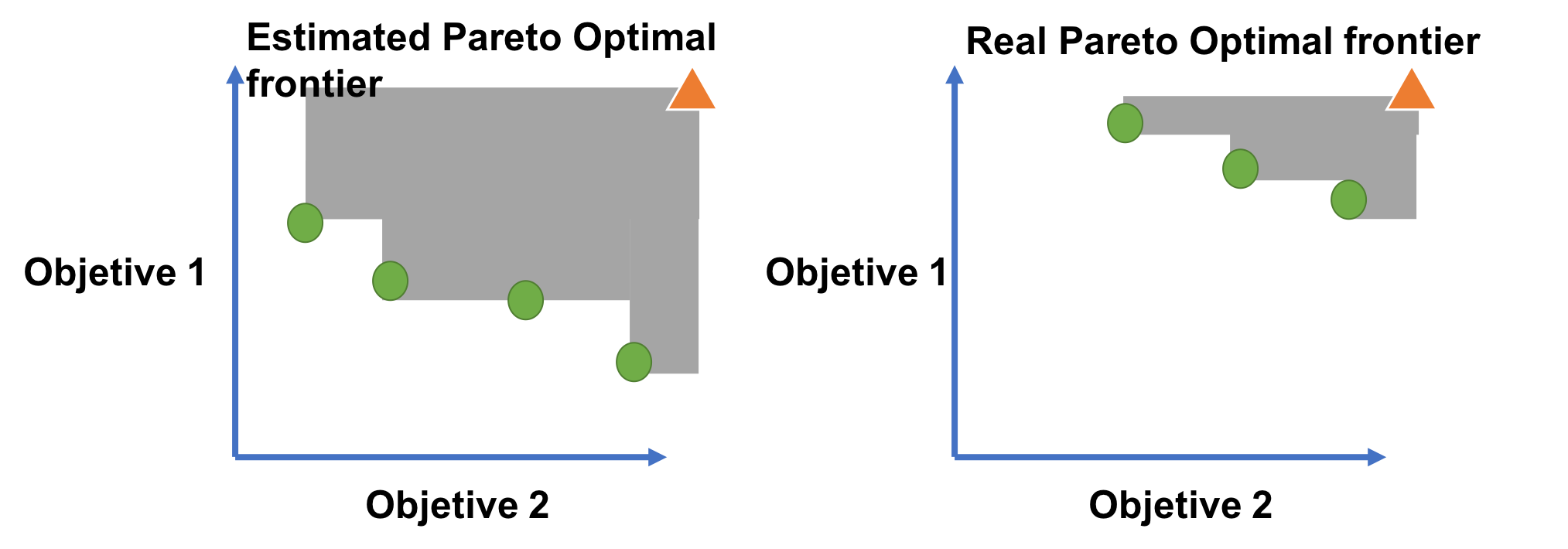}
        \caption{An example of the hypervolume between a reference point and the estimated Pareto optimal frontier (left) and the real Pareto optimal frontier (right).}
        \label{smetric}
    \end{figure}

The metrics employed to study the quality of the Pareto optimal frontier obtained by the different configurations are:
\begin{itemize}
	\item The S metric \cite{knowles2002metrics}. This metric takes a reference point above the real Pareto optimal frontier provided by the researcher, and it calculates the hypervolume between the estimated Pareto optimal frontier and the reference point. The closest the estimated Pareto optimal frontier is to the real frontier, the lower the volume will be between the estimated frontier and the reference point. Figure \ref{smetric} shows the hypervolume between a reference point and the estimated Pareto optimal frontier (left), and the hypervolume between the reference point and the real frontier. As it can be observed, the closest the estimated frontier is to the real frontier, the lower the hypervolume will be between the frontier and the reference point, with the lowest being when the estimated Pareto optimal frontier is equal to the real one. In the experiments, we take (1.0, 1.0) as the reference point.
    \item The maximum fitness found for the students.
    \item The maximum fitness found for the supervisors.
\end{itemize}    
In order to decide on the best set of parameters for the mutation operator, we followed a grid search strategy on all the possible combinations of $p_{mt}$ and $p_{sw}$.

\begin{figure}
\centering
    \includegraphics[width=0.45\linewidth]{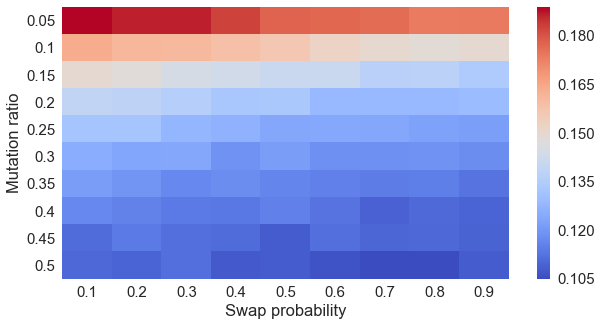}
    \includegraphics[width=0.45\linewidth]{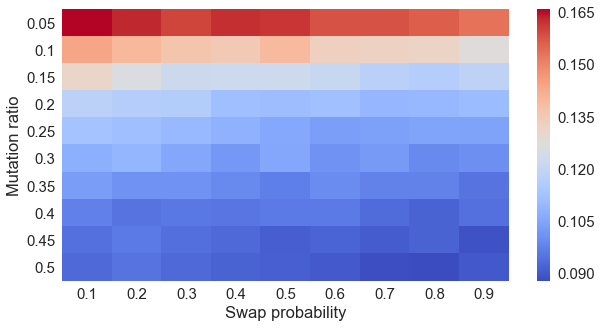}
    \includegraphics[width=0.45\linewidth]{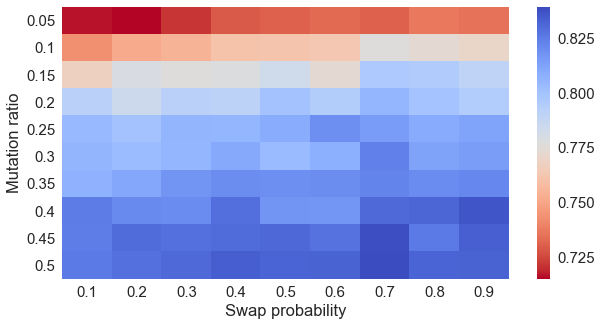}
    \caption{Average maximum fitness for the students (left), for the supervisors (right), and the average hypervolume (bottom) on the different combinations of the mutation rate ($p_{mt}$) and the probability of swapping supervisors ($p_{sw}$) when mutating a gene}
    \label{fig:hm}
\end{figure}

The results of this experiment can be seen in Figure \ref{fig:hm}. The left heatmap shows the average maximum fitness for the students, while the right heatmap contains the average maximum fitness for the supervisors. Finally, the bottom heatmap contains the average hypervolume defined by the reference point and the estimated Pareto optimal frontiers. All of the three heatmaps show a similar trend. In general, the mutation operator is more effective when a small ratio of the genes are mutated (i.e., $p_{mt}=0.05$), thus obtaining a mutated solution on the close neighborhood of the parent. Moreover, apart from remaining in the close neighborhood of the parent, the GA benefits from transferring students from one supervisor to another rather than swapping students between supervisors. As a consequence, the best values for $p_{sw}$ tend to be low and between 0.1 and 0.2. Another way to interpret this result is that the mutation operator is more suited to the problem when it explores new allocation structures instead of remaining on the same allocation structure. This result is important, as both proposed crossover operators do not explore solutions with a new allocation structure and, therefore, the goal of the mutation operator will be that of introducing new allocation structures into the population.

\subsection{Studying the Hopcroft-Karp and Greedy structural preservation crossover}


As part of the design of our genetic proposal, we have proposed two new crossover operators that are specifically designed for the problem of allocating students to supervisors. Next, we study some of the practical properties of those operators. More specifically, we will focus on studying the experimental temporal cost of both crossover operators, as well as identifying the ratio of new genetic material introduced by the \textit{greedy structural preservation} crossover operator.
 
In Section \ref{sec:xover} we studied the worst case temporal complexity of both genetic operators, with the \textit{greedy structural preservation} operator having a complexity of $\mathcal{O}(|\mathcal{S}|)$, and the \textit{Hopcroft-Karp} operator having a worst case complexity of $\mathcal{O}(|\mathcal{S}| \sqrt[]{|\mathcal{S}|})$ and an expected average complexity of $\Theta( |\mathcal{S}| log |\mathcal{S}| )$. In the following experiment we study the experimental time complexity of both operators and corroborate their adherence to their expected complexities. 

In this experiment, we ranged the number of students from 50 to 500 with steps of 50. The number of supervisors was set to one tenth of the number of students. The minimum and maximum supervision quotas of supervisors were set as described in the previous experiment. We generated one problem for each number of students. For each problem, we generated 1000 pairs of solutions that would become parents for the crossover operations. Then, for each number of students, we measured the average time taken by both crossover operators over the available pairs of solutions.

\begin{figure}
	\includegraphics[width=0.495\linewidth]{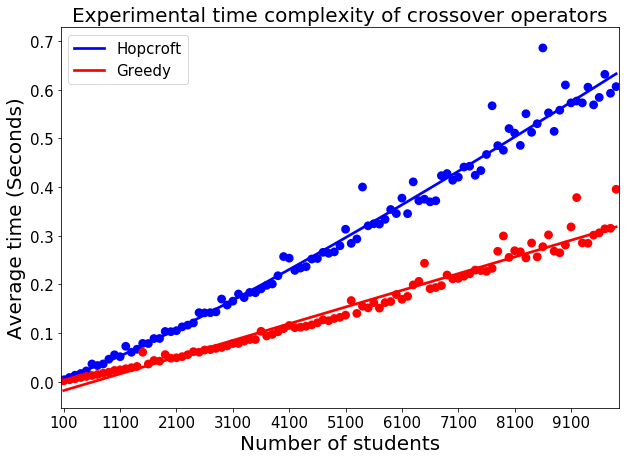}
    \includegraphics[width=0.495\linewidth]{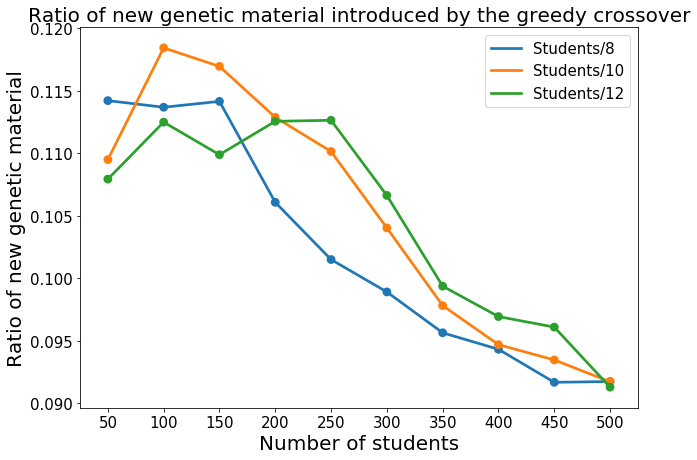}
    \caption{Average time spent by the the \textit{Hopcroft-Karp} and the \textit{greedy structural preservation} crossover operators (left), and the average ratio of new genetic material (right) introduced by the \textit{greedy structural preservation} operator with different number of students supervisors. }
    \label{img:exp_xover}
\end{figure}
The results of this experiment can be found in the left graph in Figure \ref{img:exp_xover}. The graph shows the average time spent by the \textit{Hopcroft-Karp} (blue dots) and the \textit{greedy structural preservation} (red dots) operators over allocations with different number of students. The dot markers represent the experimental data collected from the experiment, while the lines show the best fitting functions for the experimental points. One can observe that the \textit{greedy structural preservation} operator is generally faster than the \textit{Hopcroft-Karp} operator, with the differences being greater as the number of students increases. This results is aligned with our initial expectations and the suggested temporal complexity for both operators, as the \textit{Hopcroft-Karp} crossover was expected to behave at most as an $\mathcal{O}(|\mathcal{S}| log |\mathcal{S}| )$ algorithm. The best fitting function by least squares approximation for the \textit{Hopcroft-Karp} operator is a $n log n$ linearithmic function, and the best fitting function for  \textit{greedy structural preservation} operator is, as expected, a linear function. This confirms our initial hypothesis with regards to the \textit{Hopcroft-Karp} operator, with the average time being close to the case when the underlying graph is random and sparse bipartite. As a result, the operator can tackle larger problem sizes with a reasonable time, making it more applicable in a realistic context.

The next experiment that we carried out over our crossover operators has the aim of studying the ratio of new genetic material introduced by the \textit{greedy structural preservation} operator. As it was mentioned, the operator preserves the structure of one of the two parents (i.e., the number of students allocated to each supervisor) but there may be some new genetic material that is not present in any of the two parents.

For this experiment, we ranged the number of students from 50 to 500 with steps of 50, and the number of supervisors was set to be either one eighth, one tenth, or one twelfth of the number of students. Again, the minimum and maximum capacities of supervisors were set as described in the other experiments. For each combination of number of students and number of supervisors we generated a random problem. Then, for each problem we generated again 1000 pairs of solutions to act as parents for the \textit{greedy structural preservation} operator. For each combination of number of students and supervisors we measured the average ratio of new genetic material (i.e., number of new genes over the total number of genes) introduced by the operator over the 1000 crossover operations carried out. 

The results of this experiment can be found in the right graph in Figure \ref{img:exp_xover}. The first observation that can be made is that, regardless of the proportion between the number of students and supervisors, the trend appreciated is similar and so is the average ratio of new genetic material introduced in the three scenarios. The ratio of new genetic material tends to become smaller as the size of the problem becomes larger, with the highest ratios found at small number of students. Despite these ratios being higher with smaller problems, they should not be considered as disruptive with respect to the original parents. In fact, the average ratio of new genetic material in the experiments ranges from approximately 9\% to approximately 12\% of the genes. Therefore, the new genetic material introduced by the operator only explores the close neighborhood of the genetic material of both parents. If the two parents have a good fitness, one can expect that the child will yield a similar or better fitness as only a small disruption is introduced in the original genetic material.

\subsection{Optimizing the crossover operator}

Once we carried out an initial study on the behavior of the crossover operators introduced in this article, we carried out an experiment to select the best crossover operator for the problem from the ones proposed in this work and some classic and well-known crossover operators. With that goal in mind, we devise the following experiment:

\begin{itemize}
	\item We ranged the number of students from 50 to 150 with steps of 50, and we ranged the number of supervisors from 5 to 30 with steps of 5. All of the student and supervisor profiles were selected from the collected dataset. The minimum workload of supervisors $c_{j,min}$ was set to 1 student (i.e., a supervisor will advise at least one student) and the upper bound supervision quota $c_{j,max}$ of each supervisor was generated from a uniform distribution $U(4,10)$, guaranteeing that the sum of all supervisors upper bound quotas exceeded in 10, 15, or 20\% the total number of students. For each combination of number of students, number of supervisors, and upper bound supervision quotas we generated 5 different problems. This gives a total of $3 \times 6 \times 3 \times 5 = 270$ different problems.
    \item The mutation operator was set with a mutation ratio $p_{mt}=0.05$ and the probability of carrying out a swap operation in a gene to be mutated was set to $p_{sw}=0.2$. These values were found to be one of the best performing in the first experiment.
    \item We tested the performance of the \textit{Hopcroft-Karp}, the \textit{greedy structural preservation}, the \textit{uniform} \cite{syswerda89}, and the \textit{8-point}\footnote{This value of k was found to perform the best for the problems at hand} \cite{de1992formal} crossover operators. 
    \item The rest of the parameters were adjusted in the same way as defined in the experiment carried out to optimize the mutation operator.
\end{itemize}

Similarly to the first experiment, we employed the S-metric, the best fitness found for the students, and the best fitness found for the supervisors as metrics to assess the quality of the different configurations. The results of this experiment can be observed in Table \ref{tab:general_perf}. This table contains 4 sub-tables that describe the performance of the different crossover operators on the problem set.

The first subtable in Table \ref{tab:general_perf} summarizes the performance of the genetic algorithm configured with the proposed crossover operators plus the uniform and 8-point crossover operator. At a first glance, one can observe that the \textit{greedy structural preservation} operator tends to outperform the rest of crossover operators for all of the metrics. A one-sided Mann-Whitney test\footnote{$\alpha=0.05$, with the alpha values adjusted with the Bonferroni-Holm correction} comparing the performance of the aforementioned crossover operator with the individual performance of each of the other three crossover operators was carried out to assess the statistical significance of the results. The test suggests that the \textit{greedy structural preservation} operator outperforms the rest of the crossover operators for the S-Metric (i.e., the quality of the estimated Pareto frontier, to be minimized), and the best utility found for the supervisors. With regards to the best utility found for the students, the \textit{greedy structural preservation} operator was also the best performing operator, although this time we could not find statistical differences with the uniform crossover. Another interesting point that should be raised is that the \textit{Hopcroft-Karp} crossover tends to be amongst the worst performing operators from the set. Despite the similarities between the \textit{greedy structural preservation} and the \textit{Hopcroft-Karp} operator, the results suggest that the ratio of new genetic material introduced by the \textit{greedy structural preservation} crossover is beneficial for the problem at hand. In addition to this, the temporal complexity of the operator is lower than the \textit{Hopcroft-Karp} operator, making it more appropriate for this problem.

The other three subtables offer a more detailed view on the performance of the crossover operators with problems of different size. Each cell represents the performance of a given crossover operators with the problems of a given size. The performance is summarized in the form of the average over the different problems of that size, and the percentage of the problems of that size for which the crossover operator outperform the other operators. A closer look at the three subtables suggests that for the smaller problem instances (i.e., 50 students) the four crossover operators tends to perform similarly. As the problem size increases, so does the difference between the \textit{greedy structural preservation} crossover and the rest of the operators. For the larger problem instances, the proposed crossover operator is the best performing operator for all of the metrics. It should be highlighted that this is particularly true for the S-Metric and the best fitness found for the supervisors, as the operator was found to outperform the other three for 89\% and 99\% of the cases respectively. This indicates that the \textit{greedy structural preservation} operator is more suited for larger problem instances, making it the best choice overall from the studied set.
\begin{table}
\tiny
\centering
\begin{tabular}{l|l|l|l|l|}
\cline{2-5}
                                            & \multicolumn{1}{c|}{\textbf{Hopcroft}} & \multicolumn{1}{c|}{\textbf{Greedy}} & \multicolumn{1}{c|}{\textbf{Uniform}} & \multicolumn{1}{c|}{\textbf{8-point}} \\ \hline
\multicolumn{1}{|l|}{\textbf{S-Metric}}     & 0.620                                  & \textbf{0.604}                       & 0.614                                 & 0.620                                 \\ \hline
\multicolumn{1}{|l|}{\textbf{Best F. Stu}}  & 0.238                                  & \textbf{0.245}                       & \textbf{0.244}                        & 0.239                                 \\ \hline
\multicolumn{1}{|l|}{\textbf{Best F. Sup.}} & 0.230                                  & \textbf{0.242}                       & 0.231                                 & 0.227                                 \\ \hline
\end{tabular}

\vspace{0.2cm}

\begin{tabular}{|l|c|c|c|c|}
\hline
\textbf{S-Metric}  & \textbf{Hopcroft}                                    & \textbf{Greedy}                                      & \textbf{Uniform}                                     & \textbf{8-point}                                     \\ \hline
\textbf{$|\mathcal{S}|$ = 50}  & \begin{tabular}[c]{@{}c@{}}22\%\\ 0.626\end{tabular} & \begin{tabular}[c]{@{}c@{}}30\%\\ 0.624\end{tabular} & \begin{tabular}[c]{@{}c@{}}30\%\\ 0.625\end{tabular} & \begin{tabular}[c]{@{}c@{}}17\%\\ 0.626\end{tabular} \\ \hline
\textbf{$|\mathcal{S}|$ = 100} & \begin{tabular}[c]{@{}c@{}}12\%\\ 0.613\end{tabular} & \begin{tabular}[c]{@{}c@{}}62\%\\ 0.600\end{tabular} & \begin{tabular}[c]{@{}c@{}}15\%\\ 0.608\end{tabular} & \begin{tabular}[c]{@{}c@{}}10\%\\ 0.613\end{tabular} \\ \hline
\textbf{$|\mathcal{S}|$ = 150} & \begin{tabular}[c]{@{}c@{}}7\%\\ 0.621\end{tabular}  & \begin{tabular}[c]{@{}c@{}}89\%\\ 0.588\end{tabular} & \begin{tabular}[c]{@{}c@{}}4\%\\ 0.607\end{tabular}  & \begin{tabular}[c]{@{}c@{}}0\%\\ 0.620\end{tabular}  \\ \hline
\end{tabular}
\vspace{0.2cm}
\begin{tabular}{|l|c|c|c|c|}
\hline
\textbf{F. Stu} & \textbf{Hopcroft}                                    & \textbf{Greedy}                                      & \textbf{Uniform}                                     & \textbf{8-point}                                     \\ \hline
\textbf{$|\mathcal{S}|$ = 50}    & \begin{tabular}[c]{@{}c@{}}19\%\\ 0.248\end{tabular} & \begin{tabular}[c]{@{}c@{}}19\%\\ 0.248\end{tabular} & \begin{tabular}[c]{@{}c@{}}33\%\\ 0.249\end{tabular} & \begin{tabular}[c]{@{}c@{}}29\%\\ 0.248\end{tabular} \\ \hline
\textbf{$|\mathcal{S}|$ = 100}   & \begin{tabular}[c]{@{}c@{}}3\%\\ 0.240\end{tabular}  & \begin{tabular}[c]{@{}c@{}}45\%\\ 0.247\end{tabular} & \begin{tabular}[c]{@{}c@{}}44\%\\ 0.246\end{tabular} & \begin{tabular}[c]{@{}c@{}}7\%\\ 0.242\end{tabular}  \\ \hline
\textbf{$|\mathcal{S}|$ = 150}   & \begin{tabular}[c]{@{}c@{}}1\%\\ 0.226\end{tabular}  & \begin{tabular}[c]{@{}c@{}}60\%\\ 0.239\end{tabular} & \begin{tabular}[c]{@{}c@{}}38\%\\ 0.236\end{tabular} & \begin{tabular}[c]{@{}c@{}}1\%\\ 0.227\end{tabular}  \\ \hline
\end{tabular}
\vspace{0.2cm}
\begin{tabular}{|l|c|c|c|c|}
\hline
\textbf{F. Sup} & \textbf{Hopcroft}                                    & \textbf{Greedy}                                      & \textbf{Uniform}                                    & \textbf{8-point}                                    \\ \hline
\textbf{$|\mathcal{S}|$ = 50}    & \begin{tabular}[c]{@{}c@{}}20\%\\ 0.228\end{tabular} & \begin{tabular}[c]{@{}c@{}}69\%\\ 0.230\end{tabular} & \begin{tabular}[c]{@{}c@{}}5\%\\ 0.226\end{tabular} & \begin{tabular}[c]{@{}c@{}}5\%\\ 0.226\end{tabular} \\ \hline
\textbf{$|\mathcal{S}|$ = 100}   & \begin{tabular}[c]{@{}c@{}}4\%\\ 0.233\end{tabular}  & \begin{tabular}[c]{@{}c@{}}96\%\\ 0.243\end{tabular} & \begin{tabular}[c]{@{}c@{}}0\%\\ 0.232\end{tabular} & \begin{tabular}[c]{@{}c@{}}0\%\\ 0.229\end{tabular} \\ \hline
\textbf{$|\mathcal{S}|$ = 150}   & \begin{tabular}[c]{@{}c@{}}1\%\\ 0.230\end{tabular}  & \begin{tabular}[c]{@{}c@{}}99\%\\ 0.251\end{tabular} & \begin{tabular}[c]{@{}c@{}}0\%\\ 0.234\end{tabular} & \begin{tabular}[c]{@{}c@{}}0\%\\ 0.226\end{tabular} \\ \hline
\end{tabular}
\caption{ The performance of the crossover operators on the S-Metric, the best fitness found for the students, and the best fitness found for the supervisors over all of the problem sets (top), and detailed over different problem sets (middle and bottom).}
\label{tab:general_perf}
\end{table}

\subsection{Studying the optimality of the genetic algorithm}

In the previous subsection we have studied the individual performance of each of the configurable components of the GA. These studies only aimed at selecting the best possible configuration, but they did not focus on studying whether or not obtained solutions could be considered as good for the problem at hand. In this section we focus on comparing the quality of the solutions found with the optimal solution found by global optimal optimization methods. More specifically, we analyze the optimality of the best fitness found for the students, and the optimality of the best fitness found for the supervisors in the genetic algorithm. In this experiment, we focus on the largest problem instances, as metaheuristics tend to degrade their performance with the size of the problem. More specifically, the experiments were designed as follows:

\begin{itemize}
	\item The number of students was set at 150, and the number of supervisors was set at 30. All of the student and supervisor profiles were selected from the collected dataset. The minimum workload of supervisors $c_{j,min}$ was set to 1 student (i.e., a supervisor will advise at least one student) and the upper bound supervision quota $c_{j,max}$ of each supervisor was generated from a uniform distribution $U(4,10)$, guaranteeing that the sum of all supervisors upper bound supervision quotas exceeded in 10, 15, or 20\% the total number of students. For each combination of number of students, number of supervisors, and maximum upper bound we generated 5 different problems. This gives a total of $3 \times 5 = 15$ different problems.
    \item We selected the \textit{greedy structural preservation} as the crossover operator for the GA.
    \item The stop criteria was changed to keep running iterations in the GA unless the S-Metric of the estimated Pareto optimal frontier has not improved in 20 iterations. At that point, we consider that the GA has converged.
    \item Given a particular problem instance, the best possible solution for the students and for the supervisors were calculated executing two different optimization problems on \textit{BARON} \cite{baron}. \textit{BARON} is a state-of-the-art global non-convex optimization algorithm that supports constrained and pure integer optimization problems. 
    \item The rest of the parameters were adjusted in the same way as defined in the previous experiment.
\end{itemize}

\begin{table}
\tiny
\centering
\begin{tabular}{c|c|c|}
\cline{2-3}
\multicolumn{1}{l|}{}                                                                             & \multicolumn{1}{l|}{\textbf{F. Students}} & \multicolumn{1}{l|}{\textbf{F. Supervisors}} \\ \hline
\multicolumn{1}{|c|}{\textbf{\begin{tabular}[c]{@{}c@{}}Supervision capacity\\ 165\end{tabular}}} & 89.5\%                                    & 93.4\%                                      \\ \hline
\multicolumn{1}{|c|}{\textbf{\begin{tabular}[c]{@{}c@{}}Supervision capacity\\ 172\end{tabular}}} & 88.3\%                                    & 93.2\%                                      \\ \hline
\multicolumn{1}{|c|}{\textbf{\begin{tabular}[c]{@{}c@{}}Supervision capacity\\ 180\end{tabular}}} & 88.1\%                                    & 94.9\%                                      \\ \hline
\end{tabular}
\caption{Average percentage of optimality obtained by the proposed genetic algorithm for the best fitness of the students, and the best fitness of the supervisors}
\label{tab:opt}
\end{table}

\begin{figure}
\centering
	\includegraphics[width=0.325\linewidth]{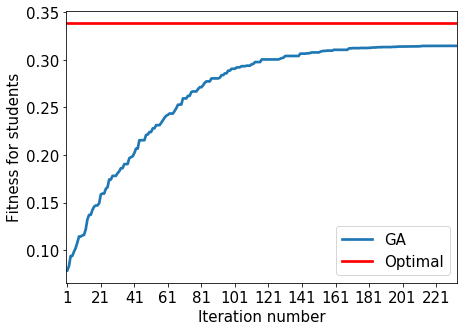}
    \includegraphics[width=0.325\linewidth]{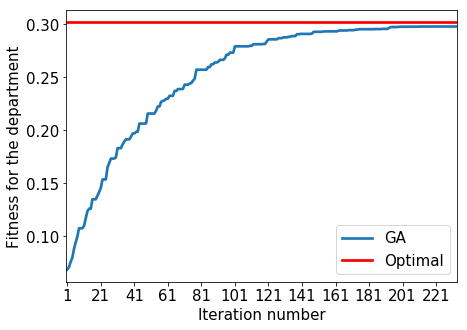}
    \includegraphics[width=0.325\linewidth]{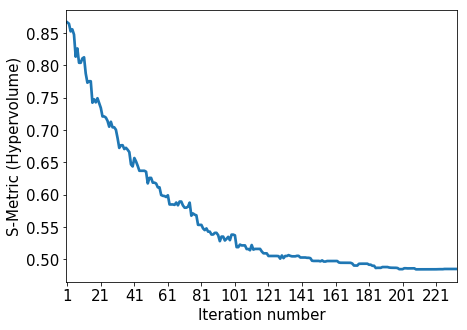}
    \caption{Convergence of the GA for an optimization problem with 180 students and 30 supervisors. The left graph shows the convergence of the best solution found for the students, the middle graph depicts the convergence of the best solution found for the supervisors, while the right graph shows the convergence of the S-Metric for the estimated Pareto optimal frontier.}
    \label{img:conv}
\end{figure}

The main results of this experiment can be found in Table \ref{tab:opt}. The table depicts the average percentage of optimality obtained for the best solution found for the students, and the best solution found for the supervisors by the GA. As it can be observed, the average percentage of optimality of the best solution for the students ranges from 88 to 89\% of the best fitness, while it ranges from 93 to 94\% for the best solution for the supervisors. These results indicate that the Pareto optimal frontier obtained by the GA contains solutions that are close to both the optimal solution for the students and the optimal solution for the supervisors. The gradual convergence of the GA for a particular problem can be observed in Figure \ref{img:conv}. As it is observable, the initial population of the GA is far from the optimal solutions (i.e., optimal solution for the students, optimal solution for the supervisors, and the distance to the reference point (1,1) in the S-Metric). As several iterations are undertaken, the GA gradually converges towards solutions that are closer to the optimal values.

Moreover, it should be highlighted that the Python implementation of the GA obtained these estimations in an average of 247 seconds, while \textit{BARON} took approximately 1020 seconds per non-linear optimization problem and only obtaining a single solution each time. Not only the estimated frontier contains solutions that are close to the optimal one for both the students and the supervisors, but these are obtained in a reasonable amount of time compared to the exact method. Nevertheless, it should be considered that our approach aims for obtaining a Pareto optimal frontier, and the exact method computes a single solution. The former is preferred from the point of view of a human decision maker with possibly uncertain preferences. In addition, the GA was capable of providing an average of 27 solutions in the estimated Pareto optimal frontier, which also provides with diversity to the human decision maker.

\section{Conclusions}
\label{ref:conclusions}
In this article we have proposed a multiobjective genetic approach for the student-supervisor allocation. This optimization problem is a subclass of the student-project allocation problem. Given the hardness of the matching problem, we have opted for a metaheuristic approach with that ability to take multiple objectives into consideration. More specifically, we take into consideration the students' preferences with regards to research/project topics, as well as the lecturers' preferences with regards to topics, which does not require the massive proposal of projects prior to the allocation, and it avoids providing explicit preferences on students as that may be regarded as a discriminatory practice. Furthermore, the genetic algorithms takes into consideration the constraints of the department in the form of lower and upper supervision quotas for lecturers, and attempts to provide a balanced workload allocation for lecturers. 

For this purpose, we have taken a Pareto optimal genetic scheme that aims to provide human decision makers with trade-off opportunities. The genetic algorithm employs a new mutation operator that can offer either explore the structure of the allocation (i.e., the number of students supervised by each lecturer) and the allocation itself. In addition, two new crossover operators have been specifically designed for the student-supervisor allocation problem: the \textit{Hopcroft-Karp} crossover operator, and the \textit{greedy structural preservation} operator. Both aim to preserve the allocation structure of one of the parents, the difference being that the \textit{Hopcroft-Karp} crossover preserves also the original genetic material from parents, while the \textit{greedy structural preservation} crossover may introduce new genetic material. The theoretical and empirical complexity of both operators has been studied, with the complexity of the former operator being linearithmic, and the complexity of the latter being linear. The genetic algorithm has been tested with real data collected from the student-supervisor allocation process at Coventry University. The results show that (i) the mutation operator benefits from giving more importance to exploring the structure of the allocation;  (ii) the \textit{greedy structural preservation} operator outperforms classic crossover operators for the problem at hand; (iii) and that the genetic algorithm is capable of providing solutions that are very close to the optimal solutions in a limited span of time, even for large problem instances.

\section*{Acknowledgements}
This work is partially supported by funds of the Faculty of Engineering and Computing at Coventry University, and funds from EU ICT-20-2015 Project SlideWiki granted by the European Commission.

\section*{References}

\bibliography{mybibfile}

\end{document}